\documentclass[journal]{IEEEtran}
\IEEEoverridecommandlockouts
\usepackage{cite}
\usepackage{amsmath,amssymb,amsfonts}
\usepackage{algorithmic}
\usepackage{graphicx}
\usepackage{textcomp}
\usepackage{color}
\usepackage{multirow}
\usepackage{xcolor}
\usepackage{subfigure}
\def\BibTeX{{\rm B\kern-.05em{\sc i\kern-.025em b}\kern-.08em
    T\kern-.1667em\lower.7ex\hbox{E}\kern-.125emX}}
\usepackage{balance}
\usepackage{float}
\usepackage[colorlinks,
            linkcolor=black,
            anchorcolor=black,
            citecolor=black]{hyperref}

\begin{document}
\title{Collaborative 3D Object Detection for Automatic Vehicle Systems via Learnable Communications}
\author{Junyong Wang, Yuan Zeng and Yi Gong
\thanks{J. Wang and Y. Gong are with Department of Electrical and Electronic Engineering. Southern University of Science and Technology (SUSTech), Shenzhen, China.
Y. Zeng is with Academy for Advanced Interdisciplinary Studies, SUSTech, Shenzhen, China. Contact e-mails:zengy3@sustech.edu.cn, gongy@sustech.edu.cn.
}
}
\maketitle
\begin{abstract}
Accurate detection of objects in 3D point clouds is a key problem in autonomous driving systems. Collaborative perception can incorporate information from spatially diverse sensors and provide significant benefits for improving the perception accuracy of autonomous driving systems. In this work, we consider that the autonomous vehicle uses local point cloud data and combines information from neighboring infrastructures through wireless links for cooperative 3D object detection. However, information sharing among vehicle and infrastructures in predefined communication schemes may result in communication congestion and/or bring limited performance improvement. To this end, we propose a novel collaborative 3D object detection framework that consists of three components: feature learning networks that map point clouds into feature maps; an efficient communication block that propagates compact and fine-grained query feature maps from vehicle to support infrastructures and optimizes attention weights between query and key to refine support feature maps; a region proposal network that fuses local feature maps and weighted support feature maps for 3D object detection. We evaluate the performance of the proposed framework using a synthetic cooperative dataset created in two complex driving scenarios: a roundabout and a T-junction. Experiment results and bandwidth usage analysis demonstrate that our approach can save communication and computation costs and significantly improve detection performance under different detection difficulties in all scenarios.
\end{abstract}
\IEEEpeerreviewmaketitle
\section{Introduction}
\label{intro}
A core component of autonomous driving vehicles or self-driving vehicles is their perceived ability to sense the surrounding environment and make decisions accordingly. 
The reliability of perception algorithms has improved significantly in the past few years due to the development of deep neural networks (DNNs) that can reason in 3D and intelligently fuse multi-sensor data, such as RGB-D images, Lidar point clouds, and GPS locations \cite{8954034, liang2018deep, 8621614}. However, precise and comprehensive perception is still a challenging task, especially when objects are heavily occluded or far away result in sparse observations. This is because the sensors equipped on a vehicle with partial observability and local viewpoints have limited sensing ability in complex driving environments. Recently, collaborative perception, via exchanging sensing information between multiple sensors using wireless communication technologies, is an emerging research trend that is receiving attention from industry and academia. Collaborative perception can enable a driving system to have a longer perception range and reduce blind spots caused by occlusions from one perspective to improve overall accuracy towards perception tasks, such as instance segmentation or object detection. Compared to local perception, cooperative perception has the advantage to augment the observation from different perspectives, as well as expanding the perception range beyond line-of-sight or field-of-view up to the boundary of autonomous vehicles\cite{7169673}.

One major challenge for cooperative perception on multiple connected agents is how to establish an effective communication mechanism to exchange information among agents, since a high bandwidth communication scheme results in network congestion and latency in the autonomous vehicle network. In this work, we formulate a cooperative 3D object detection problem where a vehicle can communicate with other agents to improve its perceptual abilities. Unlike existing works where communication protocol is predefined and/or unified, we consider a vehicle can learn to communicate with other support agents intelligently. 

To learn the communication protocol by agents themselves, most works adopt the reinforcement learning approach to multi-agent systems. They assume all agents are connected and information can be shared across all agents in a centralized manner. The key idea is learning a shared DNN to encode observations into features, then fusing features from all agents based on attention mechanism \cite{luong-etal-2015-effective, attention_is_all}, and finally decoding the fused features for perception or decision-making. A learning-based feature fusion mechanism has also been developed for autonomous vehicles. For exemple, 
in \cite{F-Cooper}, a feature fusion based cooperative perception approach was proposed for 3D object detection on autonomous vehicles. In \cite{9330564}, a cooperative spatial feature fusion approach was proposed to effectively fuse feature maps for improving 3D object detection accuracy. However, transmitting feature maps across a fully-connected graph would bring high communication costs and delays, especially when the cross-agent bandwidth is limited. We thus face the problem of designing an effective communication scheme under bandwidth constraints for cooperative 3D object detection. 

To tackle the issue, we propose an attentional communication model to enable vehicles to learn effective communication in an end-to-end deep learning manner. Inspired by recent attention mechanisms,
we designed an attention block that receives encoded local visual observation from the other connected infrastructures and determines which neighboring infrastructure should be communicated with to do cooperative detection. Each infrastructure computes a learned matching score between its local information and the received information from the vehicle. Then the autonomous vehicle selects one neighboring infrastructure to conduct the communication group. In addition, the attention model learns weights regarding feature importance for cooperative 3D object detection. The communication group dynamically changes and the entire attentional communication model is trained in a supervision manner for 3D object detection using Lidar point cloud data. This makes our approach possible for vehicles to learn coordinated strategies in dynamic wireless communication environments. In addition, since our approach decouples agent matching and data transmitting into two stages, it is suitable for large-scale vehicle networks. 

Furthermore, to evaluate the proposed approach, we design a synthetic cooperative 3D object detection dataset, named CARLA-3D by using the Car Learning to Act (CARLA) simulator \cite{CARLA}. In CARLA-3D, multiple autonomous vehicles are driving in urban environments with diverse 3D models of static objects such as roads, buildings, vegetation, traffic signs, and infrastructure, as well as dynamic objects such as vehicles and pedestrians. All models share a common scale and their sizes correspond to those of objects in the real world. We simulate two different urban driving scenarios (a roundabout and a T-junction) via varying the number of elements, including roads, sidewalks, houses, vegetation, traffic infrastructures, and locations of dynamic objects. Our experiments validate that our cooperative perception approach using the attentional communication mechanism can improve 3D object detection accuracy and reduce transmission bandwidth. The main contributions of this paper can be summarized as following:
\begin{itemize}
\item We consider a critical but under-explored task of how to effectively learn 3D object detection from point clouds via cooperative perception and present a novel framework that learns to construct communication groups under bandwidth constraints for cooperative 3D object detection in autonomous vehicle systems.    
\item To reduce communication costs, we design an attentional communication mechanism that generates an asymmetric query and adaptively fuses feature maps to balance the improvement of object detection accuracy and transmission bandwidth usage. 
\item We construct a new synthetic dataset CARLA-3D that provides Lidar point clouds for better evaluating cooperative 3D object detection with communication in autonomous vehicle systems. 
\item We analyze the effectiveness of the proposed communication mechanism on cooperative 3D object detection and show how the proposed approach outperforms local perception and centralized perception methods in terms of detection accuracy and communication costs.  
\end{itemize} 

\section{Related work}
\label{rel}
\subsection{3D object detection from point clouds}
A variety of approaches have been proposed for 3D object detection from point clouds. Similar to object detection in images, 3D object detection methods based on point clouds can be categorized into two classes: region proposal-based and single shot methods. Region proposal-based methods first propose several region proposals containing objects and then determine the category label of each proposal using extracted region-wise features. Chen et al. \cite{MV3D} introduced a sensory-fusion framework, named Multi-View 3D (MV3D), which utilizes information from Lidar point clouds and RGB images to predict oriented 3D bounding boxes. Ku et al. \cite{AVOD} proposed an Aggregate View Object Detection (AVOD) network that produces high-resolution maps from point clouds and images for autonomous driving scenarios. Qi et al. \cite{8578200} proposed Frustum PointNets for RGB-D data-based 3D object detection. In \cite{8954080}, a PointRCNN framework was proposed to generate object proposals directly from the raw point cloud and then refine the object proposals for predicting 3D bounding boxes. 

Single shot methods directly estimate class probabilities and regress 3D bounding boxes of objects using a single-stage network. Zhou et al. \cite{VoxelNet}  presented VoxelNet architecture to learn discriminative features from point cloud and detect 3D objects. Later, Yan et al. \cite{SECOND} introduced sparse convolution \cite{graham20183d} to improve the efficiency of voxel feature extraction in \cite{VoxelNet}. Sindagi et al. \cite{sindagi2019mvx} proposed MVX-Net architecture for 3D object detection via fusing image and point cloud features, which exploits multi-model information to improve detection accuracy. In \cite{PointPillars}, a 3D object detection approach named PointPillars was proposed. PointPillars first utilizes PointNet \cite{8099499} to learn feature representation of point clouds organized in vertical columns (Pillars), and then encodes the learned feature representation into a pseudo image. After that, a 2D object detection pipeline is performed to predict 3D bounding boxes with the object class. 
In contrast to doing 3D object detection from a single point of view, this paper tackles cooperative object detection problems in autonomous driving scenarios where point clouds are gathered from multiple views. We use PointPillars as a backbone to do cooperative perception due to PointPillars outperforms most fusion approaches, such as MV3D \cite{MV3D} and AVOD \cite{AVOD}.
\subsection{Collaborative 3D object detection}
Collaborative 3D object detection approaches make perception based on information from multiple sensors/agents. Chen et al. \cite{Cooper} proposed a novel neural network architecture to fuse raw point clouds collected from different positions and angles for enhancing the detection ability of autonomous driving systems. Later, an extended study on feature-level fusion schemes for collaborative perception was introduced in \cite{F-Cooper}. Hurl et al. \cite{hurl2020trupercept} proposed an end-to-end distributed perception model for cooperative perception from synthetic driving data. Guo et al. \cite{9330564} proposed a cooperative spatial feature fusion method for high-quality 3D object detection in autonomous vehicle systems. Arnold et al. \cite{9228884} introduced two cooperative 3D object detection schemes using single modality sensors. One scheme considers point clouds from multiple spatially diverse sensing points of view before detection, and the other scheme fuses the independently detected bounding boxes from multiple spatially diverse sensors. In contrast, our study tackles the problem of learning to communicate with bandwidth constraints in collaborative perception for 3D object detection. 
\subsection{Learning of communication} 
Communication is a fundamental aspect of collaborative perception, enabling connected sensors to work as a group for effective perception and decision-making in wireless sensor networks or multi-agent systems. Early works employed predefined communication protocols \cite{1580935, Tan1993MultiAgentRL} or a unified communication network \cite{singh2018learning, sukhbaatar2016learning}  instead of automatically learned communication. Recently, a few approaches have involved learning the communication of nodes/agents in multi-agent reinforcement learning. Sukhbaatar et al. \cite{Tan1993MultiAgentRL} used a neural model, named CommNet, to model the communication scheme between agents. CommNet considers full cooperation across agents, and each agent broadcasts its state to a shared communication channel and then the other agents use the integrated information for perception. Peng et al.  \cite{peng2017multiagent} proposed a bi-directional recurrent neural network (BiCNet) based communication model to integrate node-specific information from all connected nodes. Later, Jiang et al. \cite{ATOC} presented an attentional communication model to learn when and how to communicate for cooperative decision making in large-scale multi-agent systems. Liu et al. \cite{Who2com} designed a multi-stage handshake communication mechanism that learns who to communicate for reducing bandwidth usage in multi-agent cooperative perception tasks. Those tasks are built on simplistic environments where each agent observes low-dimensional 2D images. Our communication design builds upon the existing attentional communication model but focuses on cooperative perception for 3D object detection from point clouds. 
\section{Problem statement}
\begin{figure}[htbp] 
    \centering
    \includegraphics[scale = 0.20]{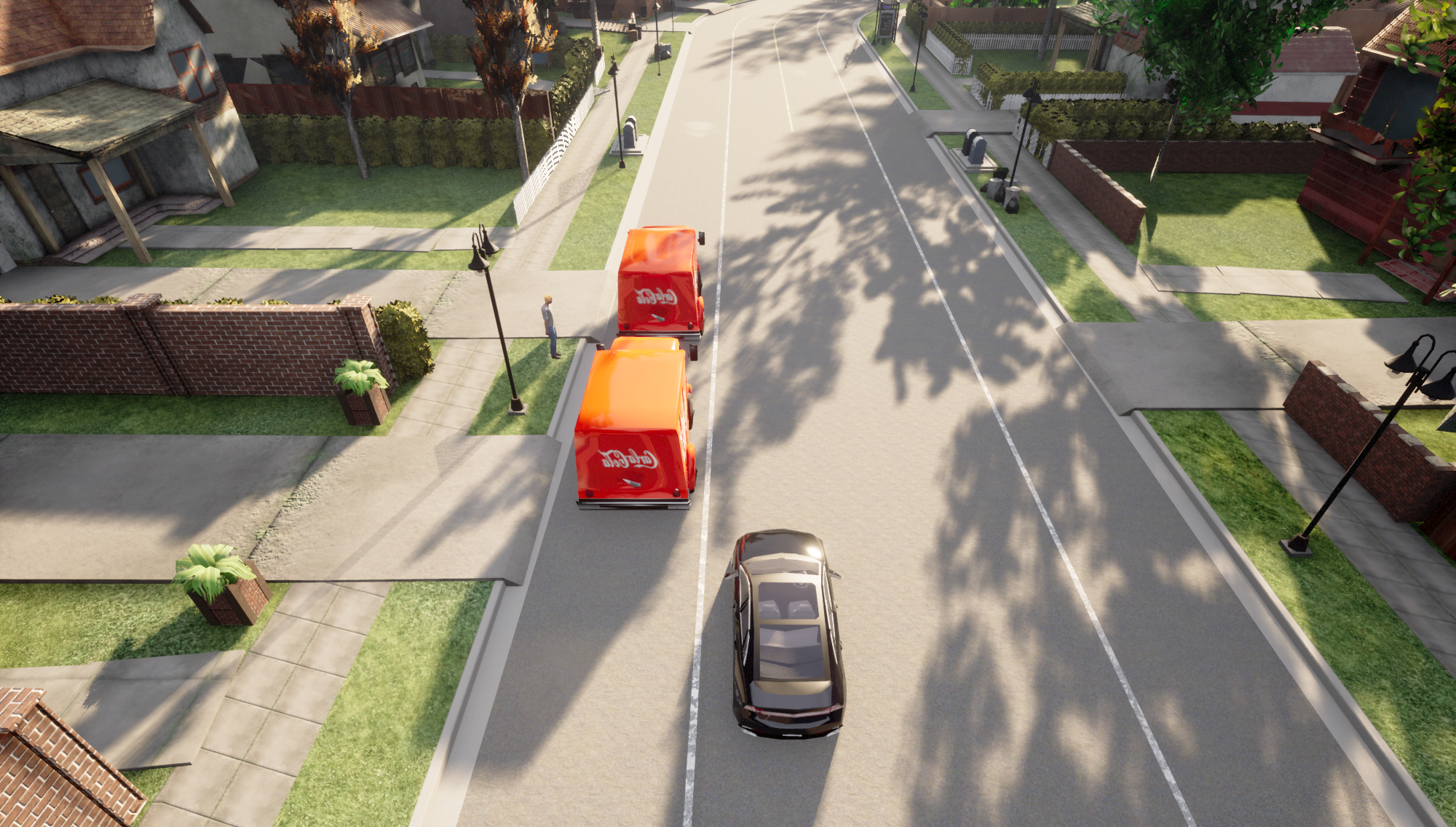}
    \caption{Illustration of an urban driving scenario where a pedestrian is heavily occluded. Due to occlusion, the autonomous vehicle has limited sensing data about the pedestrian and may ignore or give a wrong perception about the pedestrian. One possible solution is to use sensing information collected from neighboring agents, such as other vehicles or infrastructures, for cooperative detection.}
    \label{Fig_define}
\end{figure}
As shown in Fig. \ref{Fig_define}, our system model considers $N$ infrastructures. Each infrastructure is assumed to consist of a Lidar sensor and processor. Each infrastructure $i$ captures point cloud data $S_{i}$ via Lidar sensor. To benefit from cooperative perception, an autonomous vehicle $v$ is also assumed to be equipped with a wireless reception system for communication and have a local processor for perception and control, e.g., path or trajectory design. We assume that vehicles and neighboring infrastructures can exchange information through wireless links. We consider each agent (vehicle or infrastructure) has prior information about its pose, including position and orientation. For local 3D object detection, the autonomous vehicle can get local detection results $O_{v}=\mathcal{D}_{v}(\mathcal{E}_{v}(S_{v}))$ using its local point cloud data $S_{v}$ and a local processor, where $\mathcal{E}_{v}$ is a feature learning network, and $\mathcal{D}_{v}$ is a region proposal network. However, local detection may cause the vehicle to ignore the objects that are heavily occluded, such as the pedestrian in Fig. \ref{Fig_define}. To tackle the problem, the vehicle would like to integrate information from neighboring sensors for perception. In addition, sharing all information across all sensors is impractical, bandwidth and latency limitations should be considered in communication mechanisms for preventing the transmission of large amounts of or meaningless information. Our objective is thus to design a cooperative 3D object detection framework that maximizes the 3D object detection accuracy of the autonomous vehicle and minimizes the transmission bandwidth between the autonomous vehicle and neighboring infrastructures. 
\section{Cooperative 3D object detection under bandwidth constraints}
\begin{figure}[htbp] 
\centering
\includegraphics[scale = 0.46]{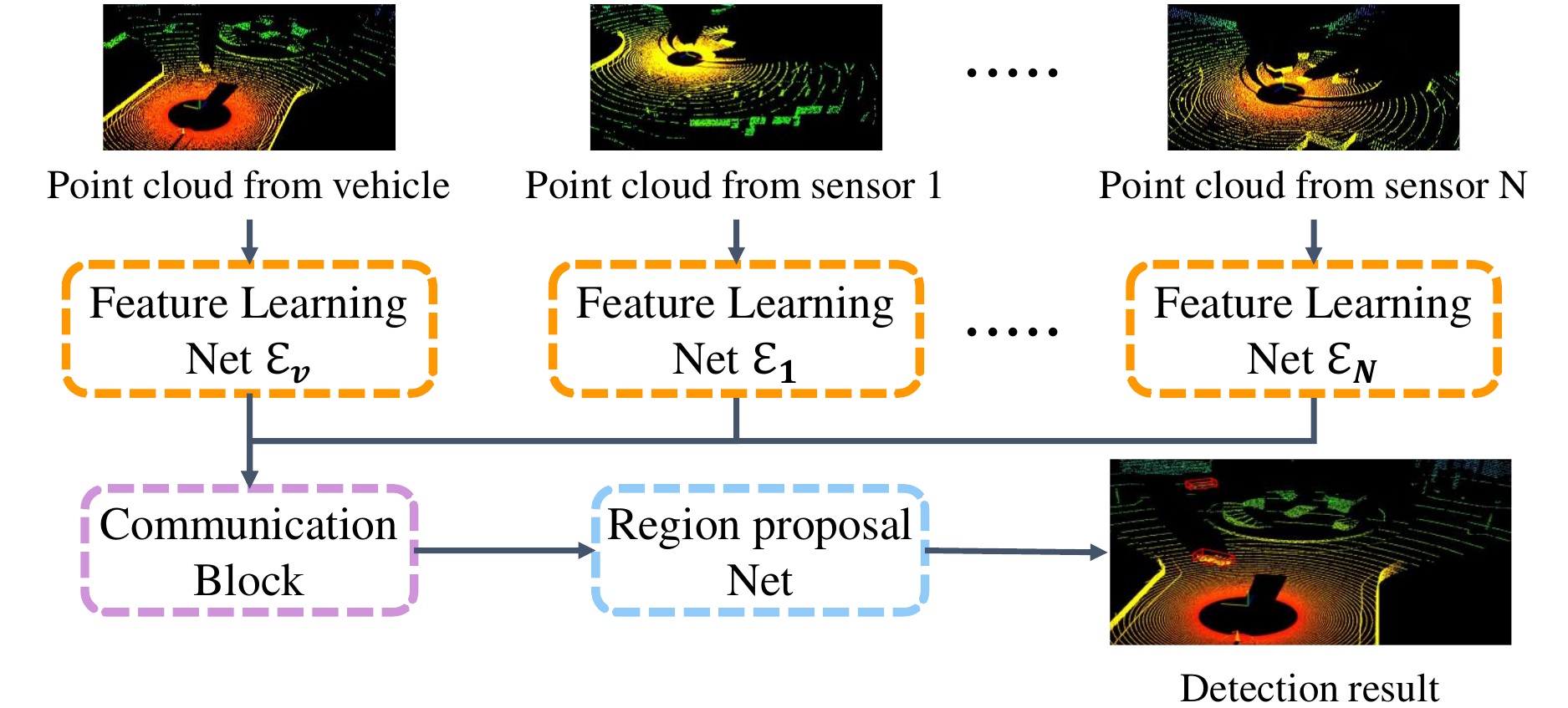}
\caption{Overview of our framework for cooperative 3D object detection in autonomous driving systems. Given point clouds from the autonomous vehicle and neighboring infrastructures, the framework first maps point clouds into feature maps locally using feature learning networks. Later, an attention-based communication block is performed to learn communication protocol using feature maps from the vehicle and infrastructures. After that, a region proposal network aims to fuse feature maps and detect objects. The framework is trained in an end-to-end manner.}
\label{Fig_framework}
\end{figure}
In this section, we introduce a framework to tackle our cooperative 3D object detection in detail. The overview of our framework is shown in Fig. \ref{Fig_framework}. Our cooperation detection framework is based on PointPillars\cite{PointPillars}, consisting of three main parts: 1) feature learning networks $\mathcal{E}$ that convert input point clouds to pseudo-images; 2) an attention-based communication block that constructs communication and cooperation between the vehicle and infrastructures; 3) a region proposal network that fuses features from the vehicle and communicated infrastructures and predicts detection results. The inputs of the proposed framework include point cloud data of the vehicle and infrastructures, and the outputs are objects label and 3D boxes predicted from the vehicle. 
\subsection{Feature learning network}
\begin{figure}[htbp] 
    \centering
    \includegraphics[scale = 0.58]{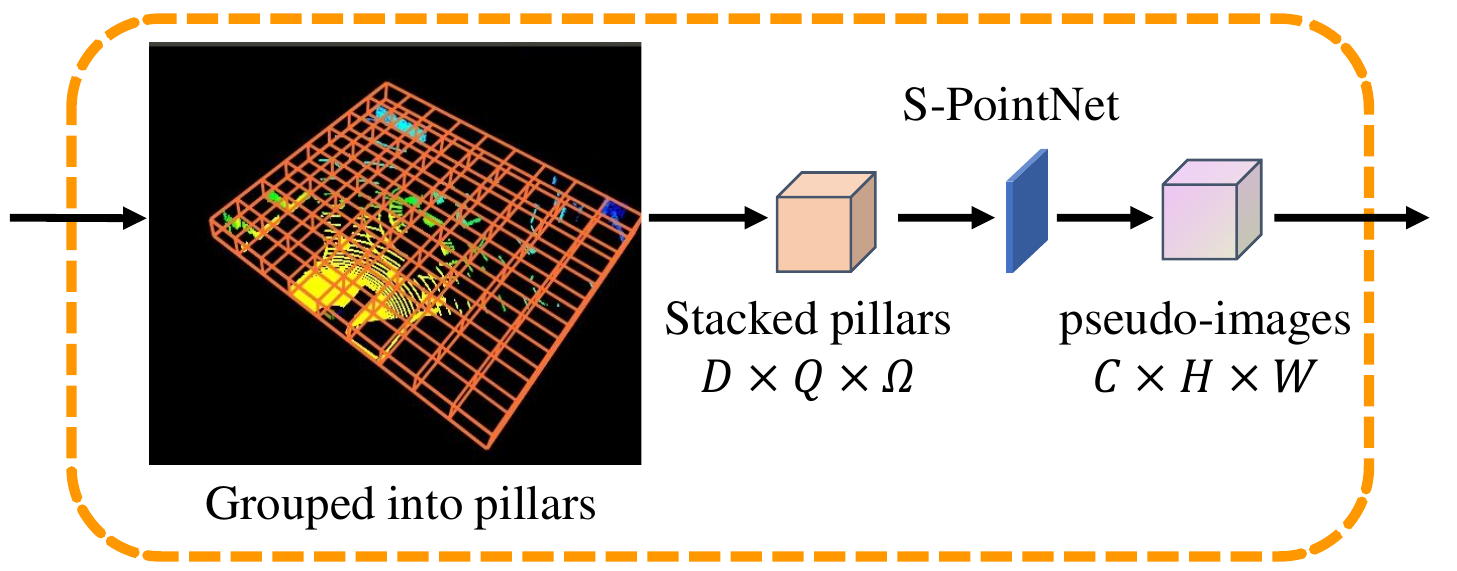}
    \caption{Architecture of the feature learning network. The input point clouds are first grouped into pillars, and then a convolution neural network is used to learn feature maps. After that, the learned feature maps are scattered back to generate a pseudo-image.}
    \label{feature_learning_net}
\end{figure}

To process communication information and detection features, infrastructures and the vehicle first convert their collected point cloud to pseudo-images, see Fig. \ref{feature_learning_net}. Similar to the first stage in Pointpillars, each infrastructure $i\in \{1,\cdots,N\}$ converts its collected point clouds $S_{i}$ to pseudo-images $\mathbf{F}_{i}=\mathcal{E}_{i}(S_{i})$ using an encoder $\mathcal{E}_{i}$, and the vehicle encodes its local point cloud data $S_{v}$ into feature maps $\mathbf{F}_{v}=\mathcal{E}_{v}(S_{v})$ using an encoder $\mathcal{E}_{v}$. The encoding process consists of three steps: creating pillars, feature learning and creating pseudo-images. Let us denote a point $p_j$ in the point cloud as $\left[x_j, y_j, z_j, r_j\right]^{T}$ with coordinates $x_j$, $y_j$, $z_j$ and reflectance $r_j$. In the first step, the point cloud is discretized into evenly spaced grids, named pillars in the x-y plane with the size of $(l_{x}, l_{y}, l_{z})$, where $l_{x}$, $l_{y}$ and $l_{z}$ denote the width, length and height, respectively. 
Then the set that all pillars are stacked defined as $\mathbf{P}=\left\{P_{k}\right\}_{k=1,\cdots,K}$. A pillar $P_k = \left\{p_{kj} = \left[ x_{kj}, y_{kj}, z_{kj}, r_{kj} \right]^{T} \in \mathbb{R}^4 \right\}_{j=1,\cdots,\Omega.}$, and contains a fixed number of points,
denoted as $\Omega$. Specifically, the $\Omega$ points are randomly selected when the number of points in a pillar is larger than $\Omega$, and zero padding is performed to get $\Omega$ points when the pillar is with fewer points due to sparsity of point cloud. A point $p_{kj}$ in pillar $P_{k}$ is then augmented with a $9$-dimentional ($D=9$) vector $(x_{kj}, y_{kj}, z_{kj}, r_{kj}, x_{ckj}, y_{ckj}, z_{ckj}, x_{okj}, y_{okj})$, where the $c$ subscript denotes the distance to the arithmetic mean of all points in the pillar, and the $o$ subscript is the offset between this point and center of the pillar in x-axis and y-axis. For each sample of point clouds, this step creates a tensor of size $D\times{Q}\times{\Omega}$, where $Q$ is the number of non-empty pillars. Then, a simplified version of PointNet (S-PointNet) is performed followed by a max operation to get learned feature maps with size of $C\times{Q}$. S-PointNet consists of a linear layer followed by batch normalization (BN) \cite{ioffe2015batch} and Rectified Linear Unit (ReLU)\cite{nair2010rectified}. After that, the tensor is scattered back to the original pillar locations to generate a pseudo-image with size of $C\times{H}\times{W}$, where $C$, $H$, and $W$ denote channel number, length, and width, respectively.
\subsection{Attention-based communication}
\begin{figure}[htbp] 
    \centering
    \includegraphics[scale=0.48]{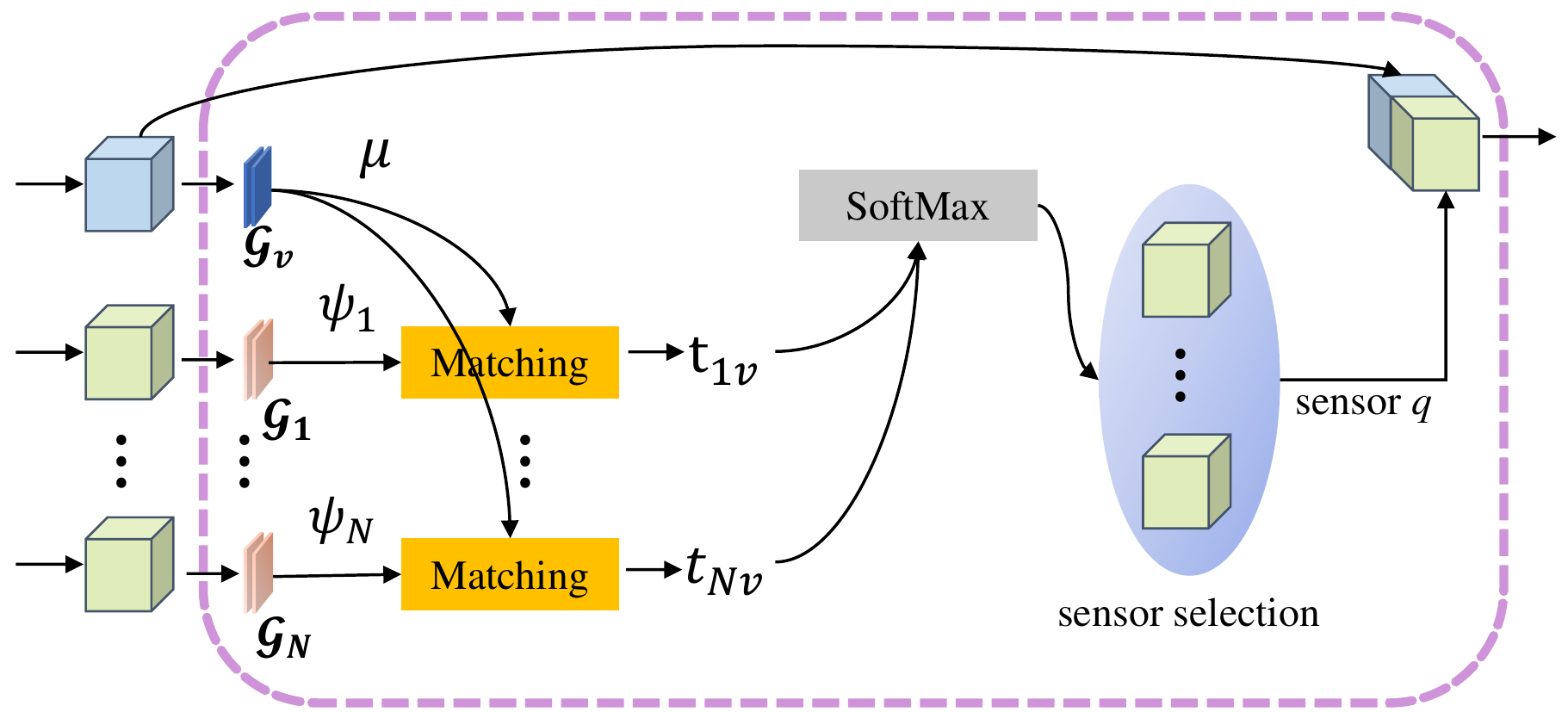}
    \caption{Illustration of the proposed attention-based communication block. It consists of three steps: 1) the vehicle computes query information and broadcasts it to all neighboring infrastructures; 2) each neighboring infrastructure computes key information and matching score between the local key information and received query information, and sends back the score to the vehicle; 3) the vehicle selects one neighboring infrastructure based on received attention scores for feature fusion.}
    \label{communication_block}
\end{figure}

In the second stage, a communication block is used to determine which neighboring infrastructure should be communicated with for improving perception accuracy. Fig. \ref{communication_block} gives an illustration of the proposed attention-based communication block. To process query information, the vehicle first encodes pseudo-images into compact query features $\mathbf{\mu}=\mathcal{G}_{v}(\mathbf{F}_{v})$ using a query network $\mathcal{G}_{v}$. Then, the vehicle broadcasts the query information to its neighboring infrastructures. After that, each infrastructure encodes pseudo-images into large-size key features $\mathbf{\psi}_{i}=\mathcal{G}_{i}(\mathbf{F}_{i})$ with a key network $\mathcal{G}_{i}$ and computes a matching score, $t_{iv}$, between the received query $\mathbf{\mu}\in{\mathbb{R}^{M_{\mu}}}$ and its key information
$\mathbf{\psi}_{i}\in{\mathbb{R}^{M_{\psi}}}$. We design $M_{\mu}\ll M_{\psi}$ to save bandwidth usage, since the query information will be broadcast from the vehicle through wireless links and the key information is used for computing matching score locally. Inspired by the attention mechanism based communication in \cite{hurl2020trupercept} and \cite{das2019tarmac}, each infrastructure computes the matching score $t_{iv}$ using the general attention mechanism in \cite{luong-etal-2015-effective}, that is
\begin{equation}
t_{iv}=\frac{\mathbf{\mu}^{T}\mathbf{W}_{a}\mathbf{\psi}_{i}}{\left\|\mathbf{\mu}^{T}\mathbf{W}_{a}\right\|\left\|\mathbf{\psi}_{i}\right\|},
\end{equation}   
where $\mathbf{W}_{a}\in{\mathbb{R}^{M_{\mu}\times{M_{\psi}}}}$ is a learnable matrix. Once all neighboring infrastructures return their matching scores back to the vehicle, the vehicle uses a softmax layer to normalize these scores into a probability distribution. For each score $t_{iv}$, a standard softmax function is used as
\begin{equation}
    \begin{aligned}
\sigma(\mathbf{t})_{i}&=\frac{\exp(t_{iv})}{\sum_{j=1}^{N}{\exp({t_{jv}})}}  \ \text{for} \ i=1, \cdots, N  \\ &\text{and } \mathbf{t}=(t_{1v}, \cdots, t_{Nv})\in{\mathbb{R}^{N}}.
    \end{aligned}
\end{equation}

The communication plays the role of integrating internal state of sensors within a group.  To process cooperative detection in region proposal network, the vehicle receives feature maps $\mathbf{F}_{q}$ from the selected infrastructure $q$, and refines the feature maps as $\mathbf{F}_{\sigma,q}$. For an infrastructure $i$, an attention refined feature map $\mathbf{F}_{\sigma,i}$ can then be generated by multiplying the attention score $\sigma(\mathbf{t})_{i}$ with the extracted feature $\mathbf{F}_{i}$, that is
\begin{equation}
\mathbf{F}_{\sigma,i}=\sigma(\mathbf{t})_{i}\otimes\mathbf{F}_{i}.
\end{equation}
Then, the vehicle concatenates $\mathbf{F}_{v}$ and $\mathbf{F}_{\sigma,i}$ to generate the refined feature maps $\mathbf{F}_{v}^{'}$, which is given by $\mathbf{F}_{v}^{'}=[\mathbf{F}_{v}, \mathbf{F}_{\sigma,q}]$. The infrastructure $q$ is searched as 
\begin{equation}
q=\arg\max_{i}t_{iv}.
\end{equation}
\subsection{Region proposal network}
\label{RPN}
\begin{figure}[htbp] 
    \centering
    \includegraphics[scale=0.48]{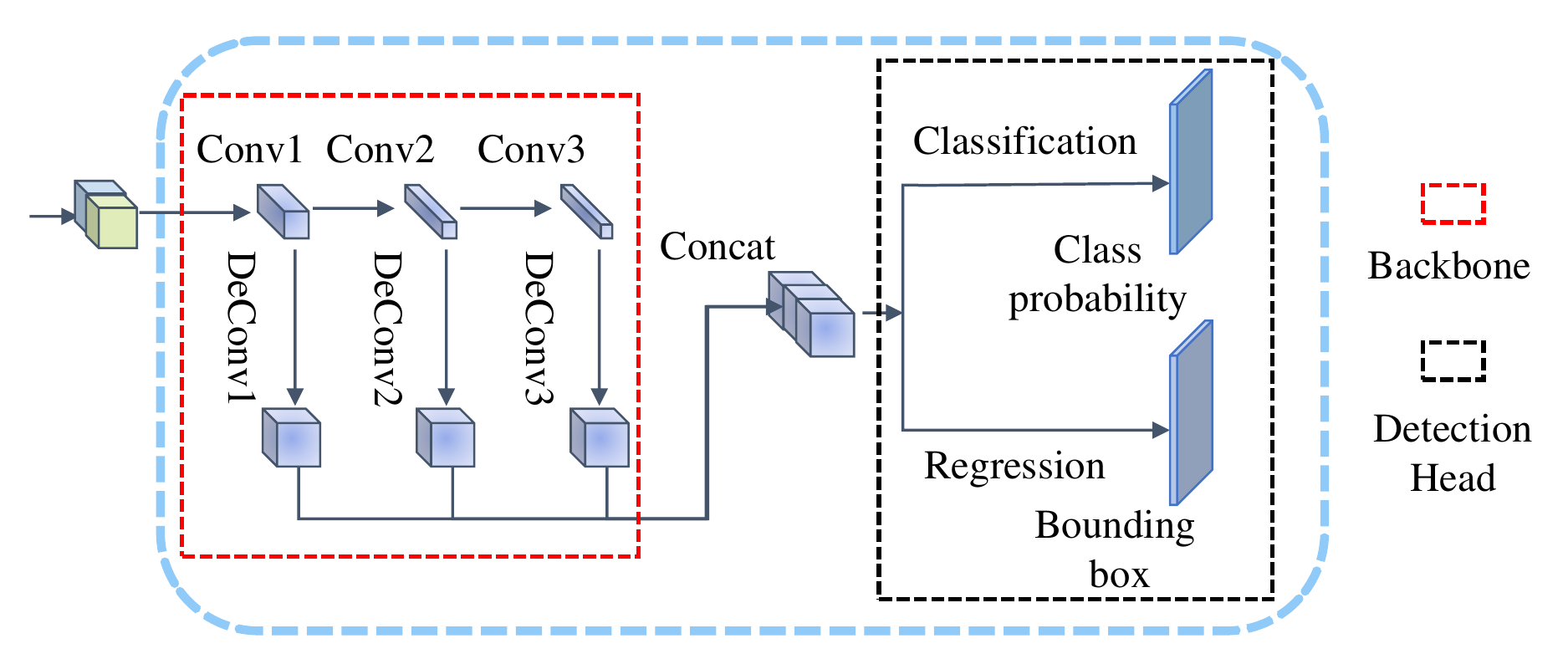}
    \caption{Region proposal network. The backbone part aims to generate multi-scale representation of the input features, and the detection head part is used to estimate a probability score and a bounding box of the proposed region.}
    \label{Decoder}
\end{figure}
In the final stage, a region proposal network $\mathcal{D}$, consisting of a backbone followed by a detection head, is performed to classify objects and predict bounding boxes. The objective of the backbone is to map the refined feature maps $\mathbf{F}_{v}^{'}$ into multi-scale representation and the detection head is then performed to detect objects as $O_{v}=\mathcal{D}_{v}(\mathbf{F}_{v}^{'})$. Inspired by the region proposal network used in Pointpillars, our backbone follows a CNN architecture similar to the one used in \cite{VoxelNet}. It processes features at three different spatial resolutions. After that, we use the Single Shot Detector Head \cite{liu2016ssd} for object classification and bounding boxes prediction. As shown in Fig. \ref{Decoder}, the backbone consists of three blocks of fully connected layers. The input feature maps $\mathbf{F}_{v}^{'}$ is down-sampled by half via convolution block with a stride size of 2. Each convolution layer is followed by BN \cite{ioffe2015batch} and ReLU\cite{nair2010rectified}. Next, we merge feature maps from the 3 different convolutional-based blocks to construct a high-resolution feature map. Since the size of feature maps from different convolutional layers are ordinarily different, we use a transposed 2D convolutional layer\cite{dumoulin2016guide} to up-sample each feature map to the size of the input features $\mathbf{F}_{v}^{'}$. After that, the high resolution feature map will be passed through two $1\times{1}$ convolutional layer to get a probability score $\varrho\in[0,1]$ and a predicted bounding box $A=(x^{a}, y^{a}, z^{a}, w^{a}, l^{a}, h^{a}, \theta^{a})$, where $(x^{a}, y^{a}, z^{a})$ denotes the center of the predicted bounding box and $(w^{a}, l^{a}, h^{a}, \theta^{a})$ indicates the width, length, height and rotation angle, respectively. We present the architecture details of the used region proposal network in Table \ref{Decoder_architecture}.
\begin{table}[htbp]
\setlength{\tabcolsep}{1.5mm}
\centering
\caption{Region proposal network architecture}
\label{Decoder_architecture}
\begin{tabular}{|c|c|c|c|c|c|}
\hline
Block                           & Layer & Filter size & Channels & Stride & Padding \\ \hline
\multirow{4}{*}{Conv1}          & Conv2d      & $3 \times 3$     & 128       & 2      & 1           \\
                                & Conv2d      & $3 \times 3$     & 128       & 1      & 1           \\
                                & Conv2d      & $3 \times 3$     & 128       & 1      & 1           \\
                                & Conv2d      & $3 \times 3$     & 128       & 1      & 1           \\ \hline
\multirow{6}{*}{Conv2}          & Conv2d      & $3 \times 3$     & 256       & 2      & 1           \\
                                & Conv2d      & $3 \times 3$     & 256       & 1      & 1           \\
                                & Conv2d      & $3 \times 3$     & 256       & 1      & 1           \\
                                & Conv2d      & $3 \times 3$     & 256       & 1      & 1           \\
                                & Conv2d      & $3 \times 3$     & 256      & 1      & 1           \\
                                & Conv2d      & $3 \times 3$     & 256       & 1      & 1           \\ \hline
\multirow{6}{*}{Conv3}          & Conv2d      & $3 \times 3$     & 512       & 2      & 1           \\
                                & Conv2d      & $3 \times 3$     & 512       & 1      & 1           \\
                                & Conv2d      & $3 \times 3$     & 512       & 1      & 1           \\
                                & Conv2d      & $3 \times 3$     & 512       & 1      & 1           \\
                                & Conv2d      & $3 \times 3$     & 512       & 1      & 1           \\
                                & Conv2d      & $3 \times 3$     & 512       & 1      & 1           \\ \hline
DeConv1                         & Deconv2D    & $1 \times 1$         & 256       & 1      & 0           \\ \hline
DeConv2                         & Deconv2D    & $2 \times 2$       & 256       & 2      & 0             \\ \hline
DeConv3                         & Deconv2D    & $4 \times 4$       & 256       & 4      & 0            \\ \hline
Regression & Conv2d      & $1 \times 1$       & 14      & 1      & 0        \\ \hline
Classification & Conv2d      & $1 \times 1$       & 2      & 1      & 0         \\ \hline
\end{tabular}
\end{table}
\begin{figure*}[htbp]
    \centering
    \subfigure[]{
        \includegraphics[scale=0.81]{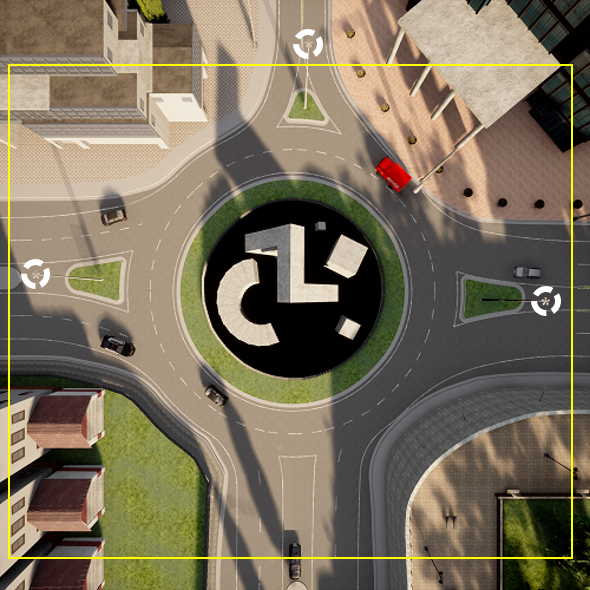}
        \label{roundabout_scene}
        }
    \subfigure[]{
        \includegraphics[scale=0.81]{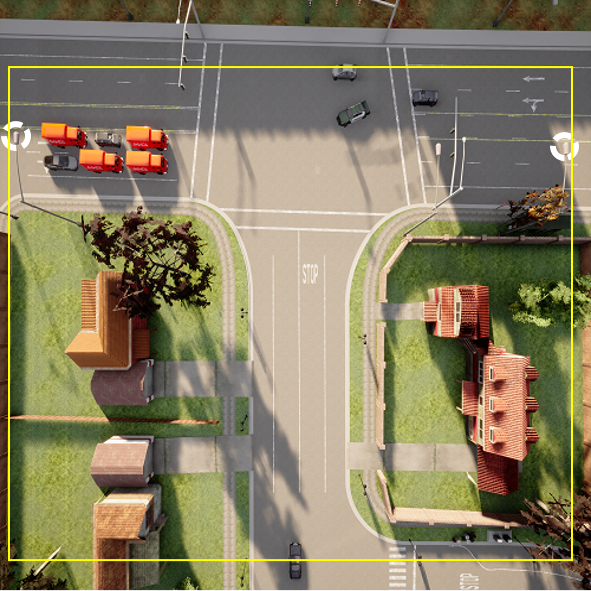}
        \label{T_junction_scene}
        }\\
        \caption{Birds Eye View of two driving scenarios. (a) roundabout; (b) T-junction. There are three Lidar sensors in roundabout and two in T-junction. The white dot circle represents the Lidar sensors, and the yellow rectangle is the detection range. The autonomous vehicle is driving from bottom to top.}
    \label{fig:scenarios}
\end{figure*}
\subsection{Training}
To train the framework and optimize the attention-based communication block in an end-to-end manner, the vehicle communicates with all neighboring infrastructures and works as a central processor during training strategy\cite{NIPS2016_c7635bfd}. Specifically, the vehicle receives scores and feature maps from all neighboring infrastructures, e.g., $(t_{iv}, \mathbf{F}_{i}), \forall{i}$, and concatenates its local feature map $\mathbf{F}_{v}$ and attention refined feature maps of all neighboring infrastructures, which is given by
\begin{equation}
\mathbf{F}_{v}^{'}=[\mathbf{F}_{v}, \sum_{i=1}^{N}\sigma(\mathbf{t})_{i}\otimes\mathbf{F}_{i}].
\end{equation}
Then, given point clouds $\mathbf{S}= \{S_{v}, S_{1}, S_{2}, \cdots, S_{N} \}$, the parameters of the framework are optimized to detect objects and predict bounding boxes $O_{v}=\left\{\varrho, A\right\}=\mathcal{D}_{v}(\mathbf{F}_{v}^{'})$ by using a joint loss that consists of classification loss $\mathcal{L}_{cls}$, localization loss $\mathcal{L}_{loc}$ and direction loss $\mathcal{L}_{dir}$. Let us denote a ground truth box as $G=(x^{gt}, y^{gt}, z^{gt}, w^{gt}, l^{gt}, h^{gt}, \theta^{gt})$, where $(x^{gt}, y^{gt}, z^{gt})$ represents the center of the ground truth box and $(w^{gt}, l^{gt}, h^{gt}, \theta^{gt})$ is the width, length, height and rotation angle, respectively. The object classification loss uses focal loss \cite{lin2017focal} defined as
\begin{equation}
\mathcal{L}_{cls}=-\eta ^{a}(1-\varrho^{a})^{\gamma}\log \varrho^{a},
\label{loss_cls}
\end{equation}
where $\varrho^{a}$ is the class probability, and $\eta ^{a}$ and $\gamma$ are hyper-parameters. The localization difference between a ground truth box and a predicted bounding box can be defined as 
\[\Delta x = \frac{x^{g t}-x^{a}}{d^{a}}, \ \Delta y = \frac{y^{g t}-y^{a}}{d^{a}}, \ \Delta z = \frac{z^{g t}-z^{a}}{h^{a}}, \]
\[\Delta w =\log \frac{w^{g t}}{w^{a}}, \ \Delta l=\log \frac{l^{g t}}{l^{a}}, \ \Delta h=\log \frac{h^{g t}}{h^{a}}, \]
\[\Delta \theta =\sin \left(\theta^{g t}-\theta^{a}\right),\]
with $d^{a}=\sqrt{(w^{a})^{2}+(l^{a})^{2}}$. The location loss between predicted boxes and ground truth boxes is defined as 
\begin{equation}
\mathcal{L}_{loc}=\sum_{b\in{(A, G)}}\text{Smooth}_{L1}(\Delta b),
\end{equation}
where $\text{Smooth}_{L1}(x)$\cite{girshick2015fast} is defined as 
\begin{equation}
 \text{Smooth}_{L1}(x) = \begin{cases}
     0.5 x^2             & \text{if } \vert x \vert < 1\\
     \vert x \vert - 0.5 & \text{otherwise}
  \end{cases}
\end{equation}
Since the localization loss cannot distinguish flipped boxes, a softmax classification loss $\mathcal{L}_{dir}$ is used to classify the boxes on discretized directions \cite{SECOND}. We generate the direction classification targets as followed:
if the $\theta^{gt}$ is higher than zero, the result is positive; otherwise, it is negative. 
The direction loss $\mathcal{L}_{dir}$ is defined as 
\begin{equation}
\mathcal{L}_{dir} = -\vartheta *log(\hat{\vartheta }) + (1 - \vartheta )*log(1 - \hat{\vartheta }),
\end{equation}
where $\vartheta $ is the heading of ground truth and $\vartheta \in \{0, 1\}$, $\hat{\vartheta}$ is the estimated heading.
The total loss function is then given by
\begin{equation}
\mathcal{L}=\frac{1}{N_{pos}}(\beta_{cls}\mathcal{L}_{cls}+\beta_{loc}\mathcal{L}_{loc}+\beta_{dir}\mathcal{L}_{dir}),
\end{equation}
where $N_{pos}$ is the number of positive anchors, and $\beta_{cls}$, $\beta_{loc}$ and $\beta_{dir}$ are scales.
\section{Experiments}
\subsection{Dataset}
\begin{figure*}[htbp] 
    \centering
    \subfigure[From vehicle in roundabout]{
    \includegraphics[scale=0.31]{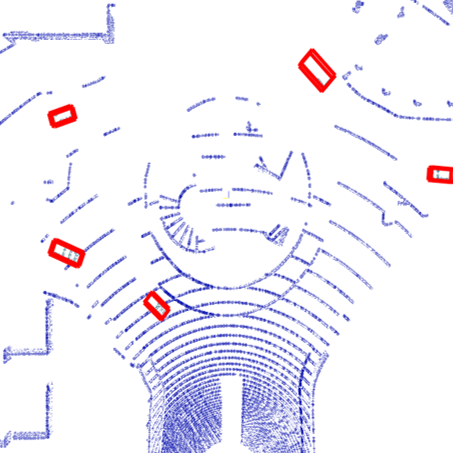}
    \label{pc_vehicle_roundabout}
    }
    \subfigure[From sensor1 in roundabout]{
    \includegraphics[scale=0.31]{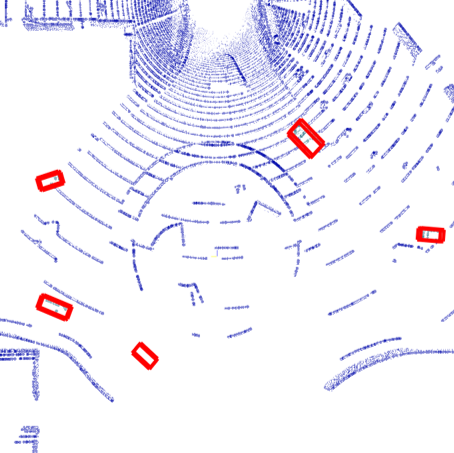}
    \label{pc_sensor1_roundabout}
    }
    \subfigure[From sensor2 in roundabout]{
    \includegraphics[scale=0.31]{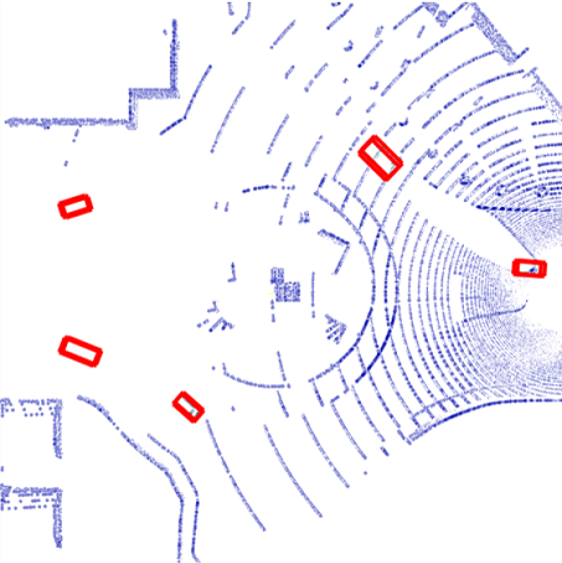}
    \label{pc_sensor2_roundabout}
    }
    \subfigure[From sensor3 in roundabout]{
    \includegraphics[scale=0.31]{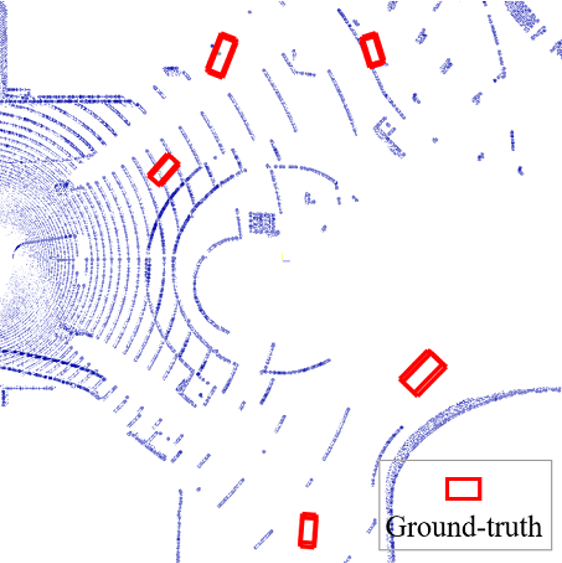}
    \label{pc_sensor3_roundabout}
    }
    \caption{Birds Eye View of point clouds data gathered from the vehicle and infrastructures in roundabout scenario given in Fig. \ref{roundabout_scene}.}
    \label{roundabout_pc} 
\end{figure*}
To evaluate the effectiveness of the proposed model for cooperative perception in autonomous vehicle systems, we use an open-source urban driving simulator Car Learning to Act (CARLA) \cite{CARLA} to generate a novel cooperative dataset.
CARLA provides open digital assets, including urban layouts, buildings, vehicles, street infrastructures, and supports flexible specification of sensor suites, environmental conditions, full control of all static and dynamic actors, and maps generation. CARLA enables simulation of complex driving scenarios as well as datasets, including Lidar data and camera data, for training and evaluation of autonomous driving systems. We design two driving scenarios: a roundabout and a T-junction, using fixed road-side infrastructure sensors and an autonomous vehicle to generate our dataset. All infrastructures and the vehicle are equipped with Lidar sensors to capture point clouds. The roundabout scenario includes three infrastructures with Lidar sensors at 2 meters ($2m$) mounting posts placed at the intersection. The T-junction scenario uses two infrastructures with Lidar sensors mounted on $2m$ high posts. All sensors are placed to fully cover the driving scenarios, as illustrated in Fig. \ref{fig:scenarios}. We capture 1788 frames data in roundabout scenario and 1610 frames data in T-junction. Each frame contains point clouds from the vehicle and infrastructure sensors and an object list describing the ground-truth position, orientation, size, and class of all objects. The objects in our dataset include vehicles and pedestrians. We set the maximum number of objects at any time frame to be 60, including 10 pedestrians and 50 vehicles. The CARLA simulator designs traffic rules and internal collision avoidance mechanisms to manage the motion of the objects. In our dataset, the state of a pedestrian is running or walking, and the probability of running pedestrian is $0.8$. In addition, we treat cars and trucks as vehicles in our dataset, where the probability that a vehicle is set as a car is $0.8$. The training set, validation set, and test set are randomly selected at a ratio of $6:2:2$ in the dataset.

Similar to official object detection datasets, e.g., KITTI\cite{KITTI_benchmark}, we consider three difficulty levels of objects: ``Easy", ``Moderate" and ``Hard", which depend on the size, occlusion level, and truncation of 3D objects.
We define objects in an image with bounding box height greater than 40 pixels, occluded area less than $33\%$ and the truncated area less than $15 \%$ as ``Easy" level; objects in an image with bounding box height greater than 25 pixels, occluded area between $33\%$ and $67\%$ and truncated area less than $30 \%$ as ``Moderate" level; objects in an image with bounding box height greater than 25 pixels, occluded area greater than $67\%$ and truncated area less than $50 \%$ as ``Hard" level.

The detection range of a Lidar sensor in our dataset is defined as a rectangle of size $80.64m\times{71.68m}$. To do cooperative object detection using point clouds from Lidar sensors located in different positions and angles, each sensor needs to map its collected Lidar data into a unified coordinate system. For exemplar, the vehicle broadcasts its location information to neighboring infrastructures together with query, and then each infrastructure maps point clouds into the positions of the vehicle. The location information of a Lidar sensor equipped on the vehicle contains its GPS coordinates $C_{v}=(x_{v}, y_{v}, z_{v})$ and rotation information, including yaw angles $\alpha_{v}$ and pitch and roll angles. Given the GPS coordinates of a point cloud $j$ of an infrastructure $i$ and the GPS coordinates of the infrastructure $i$, denoted as $C_{j,i}=[x_{j,i}, y_{j,i}, z_{j,i}]$ and $C_{i}=[x_{i}, y_{i}, z_{i}]$, respectively, the point can be transformed into the vehicle's coordinate system as 
\begin{equation}
C_{j,v}=\mathbf{R}C_{j,i}^{T}+C_{i}^{T}-C_{v}^{T}, 
\end{equation}
where $\mathbf{R}$ is a rotation matrix. Since the pitch and roll angles of autonomous vehicle and infrastructure are 0, $\mathbf{R}$ is defined as
\begin{equation}
\mathbf{R}=\begin{bmatrix}
cos (\alpha_v - \alpha_i)  & -sin (\alpha_v - \alpha_i) & 0 \\
sin (\alpha_v - \alpha_i) & cos (\alpha_v - \alpha_i) & 0 \\
0 & 0  & 1
\end{bmatrix},
\end{equation}
where $\alpha_i$ is the yaw angles of the infrastructure $i$.
\begin{figure*}[t] 
    \centering
    \subfigure[From vehicle in T-junction]{
    \includegraphics[scale=0.44]{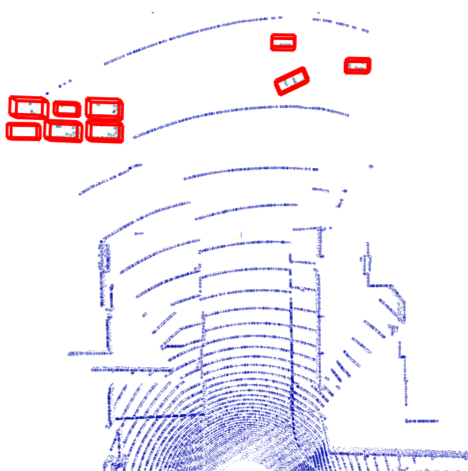}
    \label{pc_vehicle_T_junction}
    }
    \subfigure[From sensor1 in T-junction]{
    \includegraphics[scale=0.44]{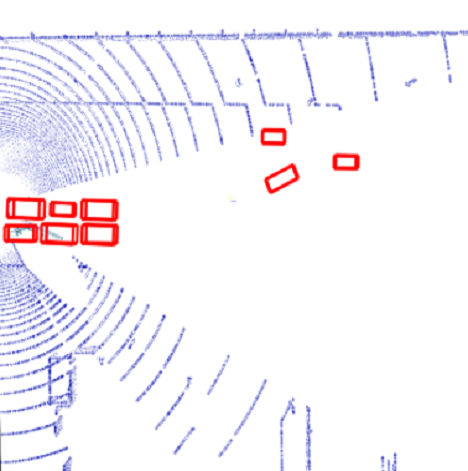}
    \label{pc_sensor1_T_junction}
    }
    \subfigure[From sensor2 in T-junction]{
    \includegraphics[scale=0.44]{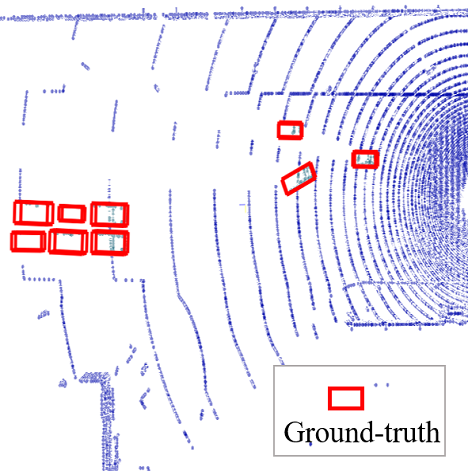}
    \label{pc_sensor2_T_junction}
    }
    \caption{Birds Eye View of point clouds data gathered from the vehicle and infrastructures in T-junction scenario given in Fig. \ref{T_junction_scene}.}
    \label{T_junction_pc} 
\end{figure*} 
\subsection{Evaluation metrics}
\begin{figure}[t]
    \centering
    \subfigure[]{
    \centering
    \includegraphics[scale=0.46]{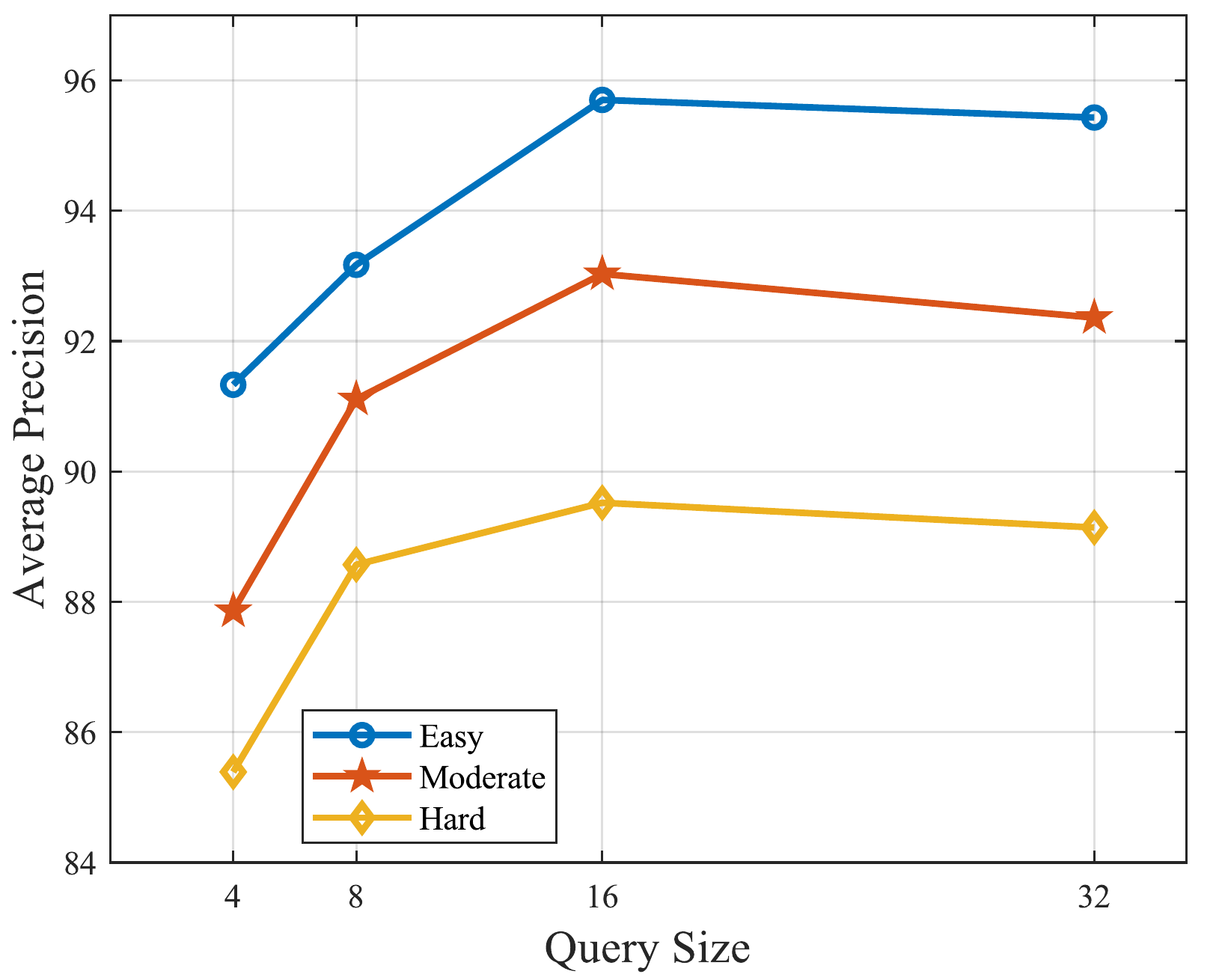}
    \label{vary_query_roundabout}
    } 
    \subfigure[]{
    \centering
    \includegraphics[scale=0.46]{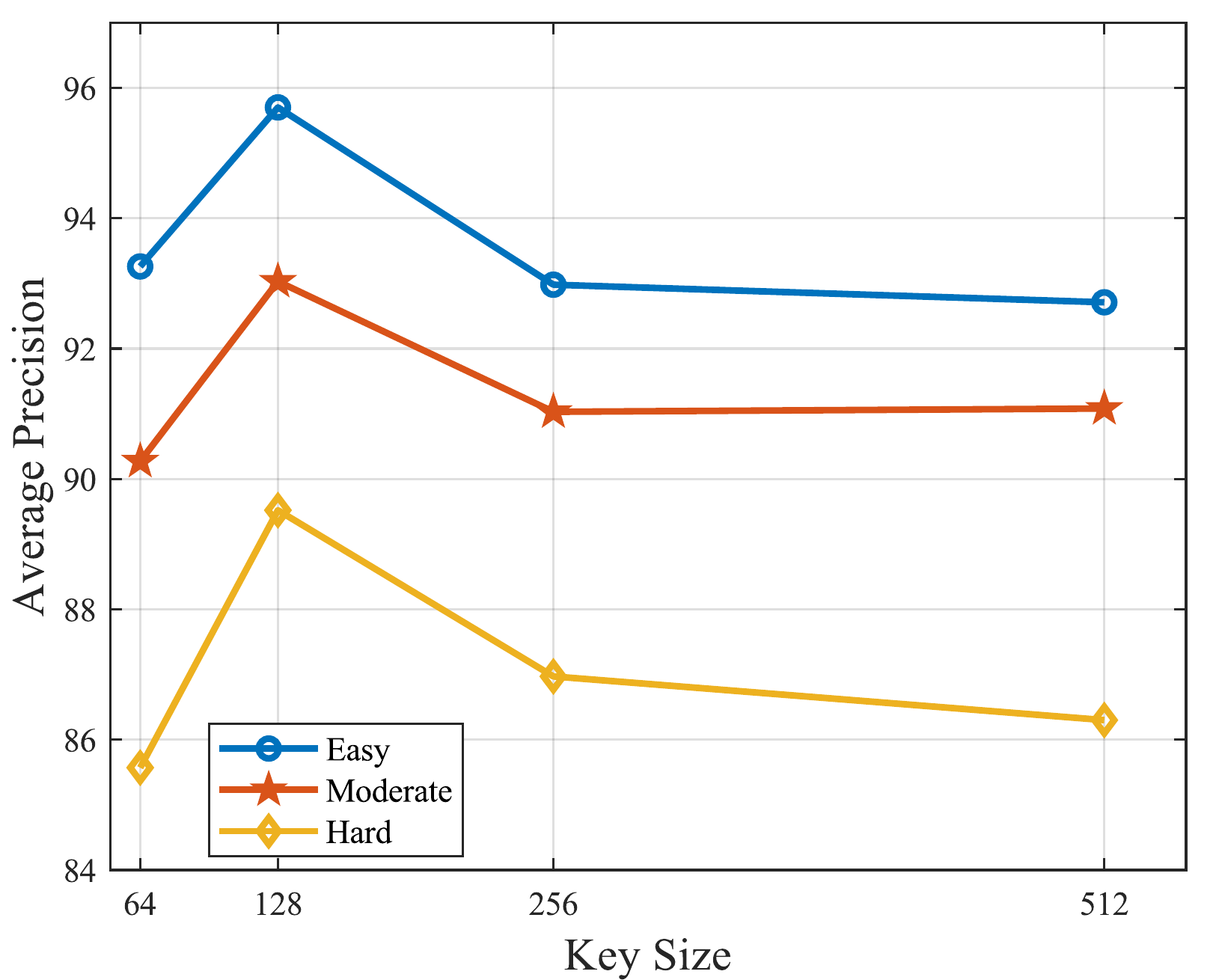}
    \label{vary_key_roundabout}
    }
    \caption{Ablation study on the size of query and keys under Easy, Moderate and Hard difficulties in roundabout scenario. (a) AP with key size of 128 versus query size from 4 to 32; (b) AP with query size of 32 versus key size from 64 to 512.}\label{query_and_key_roundabout}
\end{figure}
We use average precision (AP) \cite{VOC} as a measure to assess the detection performance of the presented framework. We also introduce an accuracy-improvement-to-bandwidth-usage (AIB) to assess communication costs. The AP is derived from precision and recall, which are single-value metrics depend on probability score and interaction over union (IoU). To do the calculation of AP for object detection, we first calculate IoU\cite{VOC}. The IoU is given by the ratio of the volume of intersection and volume of the union of the predicted bounding box $B_{pred}$ and ground truth bounding box $B_{gt}$:
\begin{equation}
IoU\left(B_{g t}, B_{pred}\right)=\frac{\operatorname{volume}\left(B_{g t} \cap B_{pred}\right)}{\operatorname{volume}\left(B_{g t} \cup B_{pred}\right)}.
\label{IoU}
\end{equation}
The IoU is a number in the range $[0, 1]$, where 0 means no overlap between $B_{g t}$ and $B_{pred}$, and 1 indicates $B_{g t}$ and $B_{pred}$ are completely overlapped. When the probability score $\varrho$ in  Eq.(\ref{loss_cls}) is greater than a threshold $\delta$ and $\operatorname{IoU}\left(B_{g t}, B_{pred}\right)$ is greater than a threshold $\sigma$, the prediction will be considered as positive, otherwise negative. The precision $e$ is defined as the ratio of the number of correct positive predictions in the prediction set, and recall $r$ is defined as the ratio of the number of correct positive predictions in ground-truth set. Then, we can calculate the corresponding AP for $K$ recall levels\cite{VOC,everingham2015the} as 
\begin{equation}
AP=\sum_{k=1}^{K}e_{\text{interp}}(r_{k+1})\left[r_{k+1}-r_{k}\right],
\end{equation}
with 
\begin{equation}
e_{\text{interp}}(r)=\max_{\tilde{r}:\tilde{r}\geq{r}}e(\tilde{r}),
\end{equation}
where $K$ is the number of predicted bounding boxes, and $e(r)$ is the precision as the function of recall $r$. The $k$th recall $r_{k}$ is computed by setting the probability score threshold $\delta$ equal to the confidence score of the $k$th estimated bounding box, sorting by the confidence score in descending order. $e_{\text{interp}}(r)$ is an interpolated precision that takes maximum precision over all recalls greater than $r$, which smooths the original precision curve $e(r)$. 

In order to evaluate and analyze the performance of our framework under limited bandwidth, we measure the bandwidth usage in terms of AIB, which is defined by 
\begin{equation}
AIB=\frac{ | \upsilon  - \upsilon ^{'} |}{B},
\end{equation}
where $\upsilon $ is the AP of the proposed communication framework, and $\upsilon ^{'}$ is the AP of the vehicle without communication and $B$ is the bandwidth usage (in Mbytes per frame) of the proposed communication framework.
\subsection{Experimental setup}
We conduct three experiments to analyze the cooperative 3D object detection performance of the proposed framework, referred as Learn2com. First, we give an ablation study on the size of query and keys. Second, we use a comparative experiment to evaluate the detection performance and bandwidth usage of our approach under Easy, Mod, and Hard difficulties. After that, we further evaluate the detection performance of our approach under far and near difficulties. We consider the following three baseline 3D object detection methods for comparison:
\begin{itemize}
\item \emph{LocVehicle}: the LocVehicle model uses the local point cloud data of the vehicle without communicating with neighboring infrastructures.
\item \emph{CombAll}: unlike the proposed approach where the vehicle adaptively selects one neighboring infrastructure for cooperative 3D object detection, the CombAll model considers that the vehicle communicates with all neighboring infrastructures for cooperative perception and each infrastructure contributes its information with the same importance factor. 
\item \emph{RandSelect}: instead of learning to select one infrastructure to communicate with, the RandSelect model randomly selects one neighboring infrastructure to communicate with for cooperative 3D object detection.
\end{itemize}
\subsection{Implementation details}
We set pillar size to $l_{x}=l_{y}=0.56m$, $l_z=4m$, and the maximum number of points per pillar $\Omega=100$. Each class anchor is described by width, length, height, $z$ center, and applied at two orientations: 0 and 90 degrees. Each anchor is matched to ground-truth and assigned to positive or negative (an object or background). 
The anchor with the highest IoU that overlaps with a ground-truth or above the positive match threshold is considered as positive, while the anchor is negative when the IoU between the anchor and all ground-truth is below the negative match threshold. The anchors with IoUs between negative match threshold and positive match threshold are ignored during training.
For cars, the anchor has width, length, and height of $(1.6m, 3.9m, 1.56m)$ with a $z$ center of $-1.78m$. For trucks, the anchor has width, length, and height of $(1.9m, 4.9m, 2.05m)$ with a $z$ center of $-1.5m$. We set positive and negative matching thresholds of all class anchors to $0.6$ and $0.45$, respectively, and set the threshold of IoU $\sigma$ to $0.7$ in evaluation metrics written as AP@IoU 0.7.

We use the same hyper-parameter settings as Pointpillars \cite{PointPillars}: $\eta ^{a}=0.25$, $\gamma=2$, $\beta_{cls}=1$, $\beta_{loc}=2$ and $\beta_{dir}=0.2$. The proposed framework is implemented in Pytorch and trained on a PC with four NVIDIA TITAN X GPUs. The models are optimized using Adam optimizer \cite{Adam}. The initial learning rate is $0.0002$ with an exponential decay factor of 0.8 and decays every 15 epochs. The training process is terminated when the validation loss converges, and the model with the best evaluation performance will be saved for test. 
\begin{figure*}[htbp]
    \centering
    {\centering
    \includegraphics[scale=0.30]{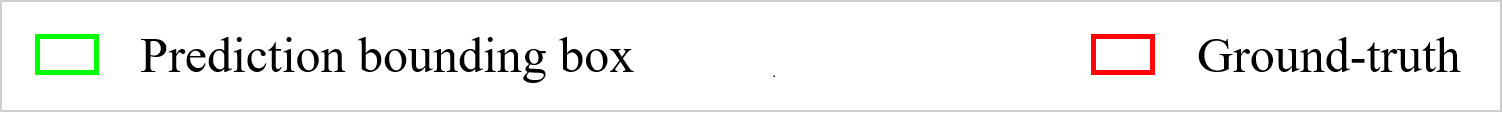}
    }
    
    \subfigure[Locvehicle]{
    \centering
    \includegraphics[scale=0.50]{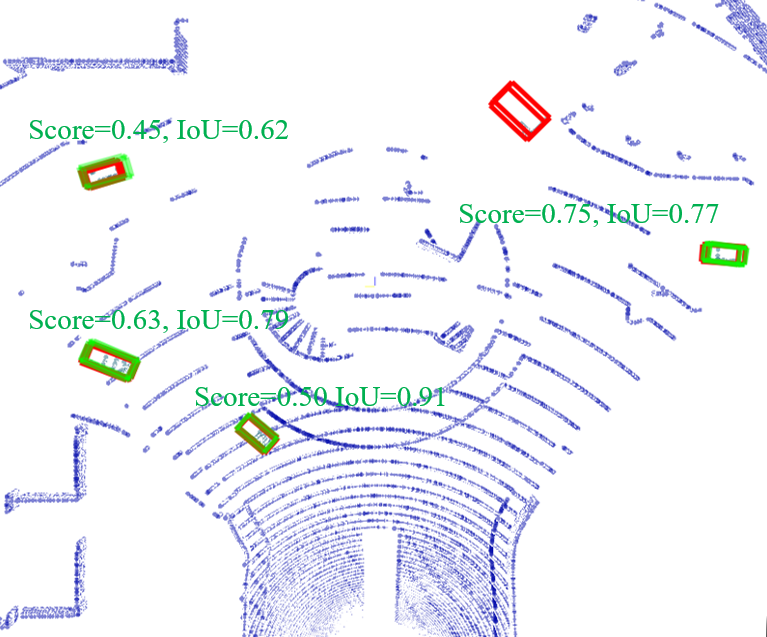}
    \label{LocaVehicle_roundabout}
    } 
    \subfigure[RandSelect]{
    \centering
    \includegraphics[scale=0.50]{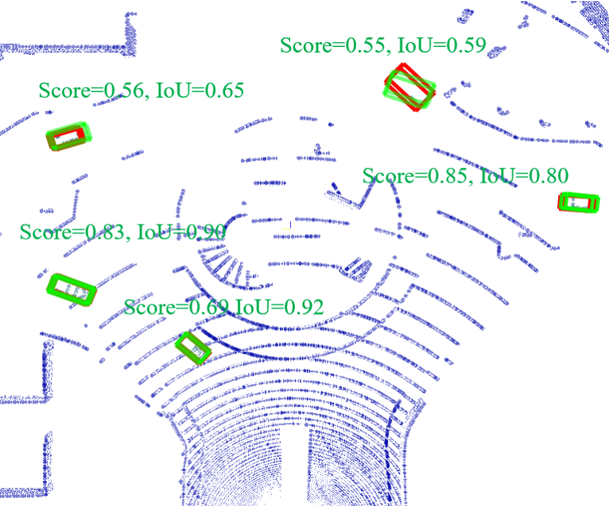}
    \label{RandSelect_roundabout}
    }
    
    \subfigure[CombAll]{
    \centering
    \includegraphics[scale=0.50]{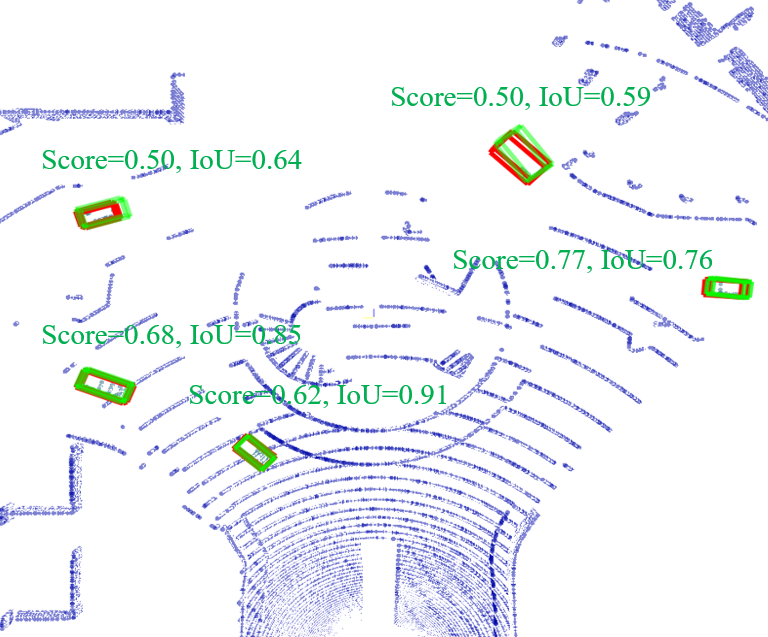}
    \label{Comball_roundabout}
    }
    \subfigure[Learn2com]{
    \centering
    \includegraphics[scale=0.50]{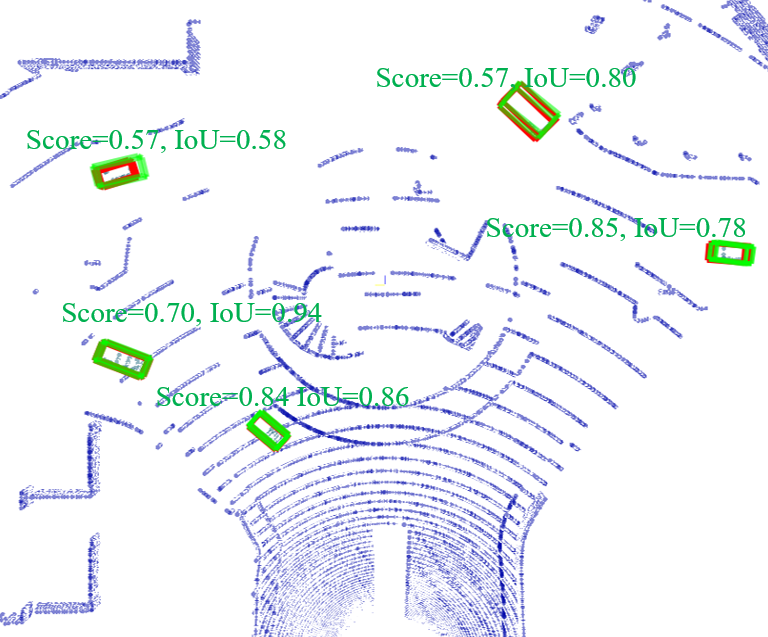}
    \label{learn2con_roundabout}
    } 

    \caption{Detection examples of different detection schemes in roundabout scenario. Our method Learn2com performs better with higher detection accuracy. 
    }
    \label{roundabout_results} 
\end{figure*}
\begin{table*}[htbp]
\setlength{\tabcolsep}{1.75mm}
\centering
\caption{Detection performance comparison on CARLA-3D under Easy, Moderate and Hard difficulties.}
\label{table:PERFORMANCE_detection_kitti}
\begin{tabular}{|c|c|cccccc|c|cccccc|}
\hline
\multirow{3}{*}{} &
   &
  \multicolumn{6}{c|}{roundabout AP@IoU 0.7} &
   &
  \multicolumn{6}{c|}{T-junction AP@IoU 0.7} \\ \cline{2-15}  &
  mAP &
  \multicolumn{3}{c|}{Car} &
  \multicolumn{3}{c|}{Truck} &
  mAP &
  \multicolumn{3}{c|}{Car} &
  \multicolumn{3}{c|}{Truck} \\ \cline{2-15} 
 &
  Moderate &
  \multicolumn{1}{c|}{Easy} &
  \multicolumn{1}{c|}{Moderate} &
  \multicolumn{1}{c|}{Hard} &
  \multicolumn{1}{c|}{Easy} &
  \multicolumn{1}{c|}{Moderate} &
  Hard &
  Moderate &
  \multicolumn{1}{c|}{Easy} &
  \multicolumn{1}{c|}{Moderate} &
  \multicolumn{1}{c|}{Hard} &
  \multicolumn{1}{c|}{Easy} &
  \multicolumn{1}{c|}{Moderate} &
  Hard \\ \hline
LocVehicle &
   88.12&
  \multicolumn{1}{c|}{89.17} &
  \multicolumn{1}{c|}{85.67} &
  \multicolumn{1}{c|}{82.11} &
  \multicolumn{1}{c|}{94.86} &
  \multicolumn{1}{c|}{90.57} &
  78.20&
   89.79&
  \multicolumn{1}{c|}{89.91} &
  \multicolumn{1}{c|}{88.42} &
  \multicolumn{1}{c|}{78.82} &
  \multicolumn{1}{c|}{92.91} &
  \multicolumn{1}{c|}{91.16} &
   78.37
   \\ \hline
RandSelect &
   91.93 &
  \multicolumn{1}{c|}{94.57} &
  \multicolumn{1}{c|}{92.03} &
  \multicolumn{1}{c|}{86.41} &
  \multicolumn{1}{c|}{96.95} &
  \multicolumn{1}{c|}{91.82} &
  83.19&
  93.03&
  \multicolumn{1}{c|}{91.28}  &
  \multicolumn{1}{c|}{89.92} &
  \multicolumn{1}{c|}{79.62} &
  \multicolumn{1}{c|}{96.78} &
  \multicolumn{1}{c|}{96.13} &
   82.74
   \\ \hline
CombALL &
  91.07&
  \multicolumn{1}{c|}{92.24} &
  \multicolumn{1}{c|}{90.16} &
  \multicolumn{1}{c|}{87.64} &
  \multicolumn{1}{c|}{96.56} &
  \multicolumn{1}{c|}{91.98} &
  84.23&
  94.28&
  \multicolumn{1}{c|}{94.87} &
  \multicolumn{1}{c|}{92.63} &
  \multicolumn{1}{c|}{82.53} &
  \multicolumn{1}{c|}{98.69} &
  \multicolumn{1}{c|}{95.93} &
   85.15
   \\ \hline
Learn2com &
  \textbf{92.68}&
  \multicolumn{1}{c|}{\textbf{95.70}} &
  \multicolumn{1}{c|}{\textbf{93.03}} &
  \multicolumn{1}{c|}{\textbf{89.52}} &
  \multicolumn{1}{c|}{\textbf{99.64}} &
  \multicolumn{1}{c|}{\textbf{92.33}} &
  \textbf{84.59}&
  \textbf{95.16}&
  \multicolumn{1}{c|}{\textbf{95.55}} &
  \multicolumn{1}{c|}{\textbf{93.40}} &
  \multicolumn{1}{c|}{\textbf{83.88}} &
  \multicolumn{1}{c|}{\textbf{99.88}} &
  \multicolumn{1}{c|}{\textbf{96.91}} &
  \textbf{86.76}
  \\ \hline
\end{tabular}
\end{table*}
\begin{figure*}[htbp]
    \centering
    \subfigure[]{
    \centering
    \includegraphics[scale=0.46]{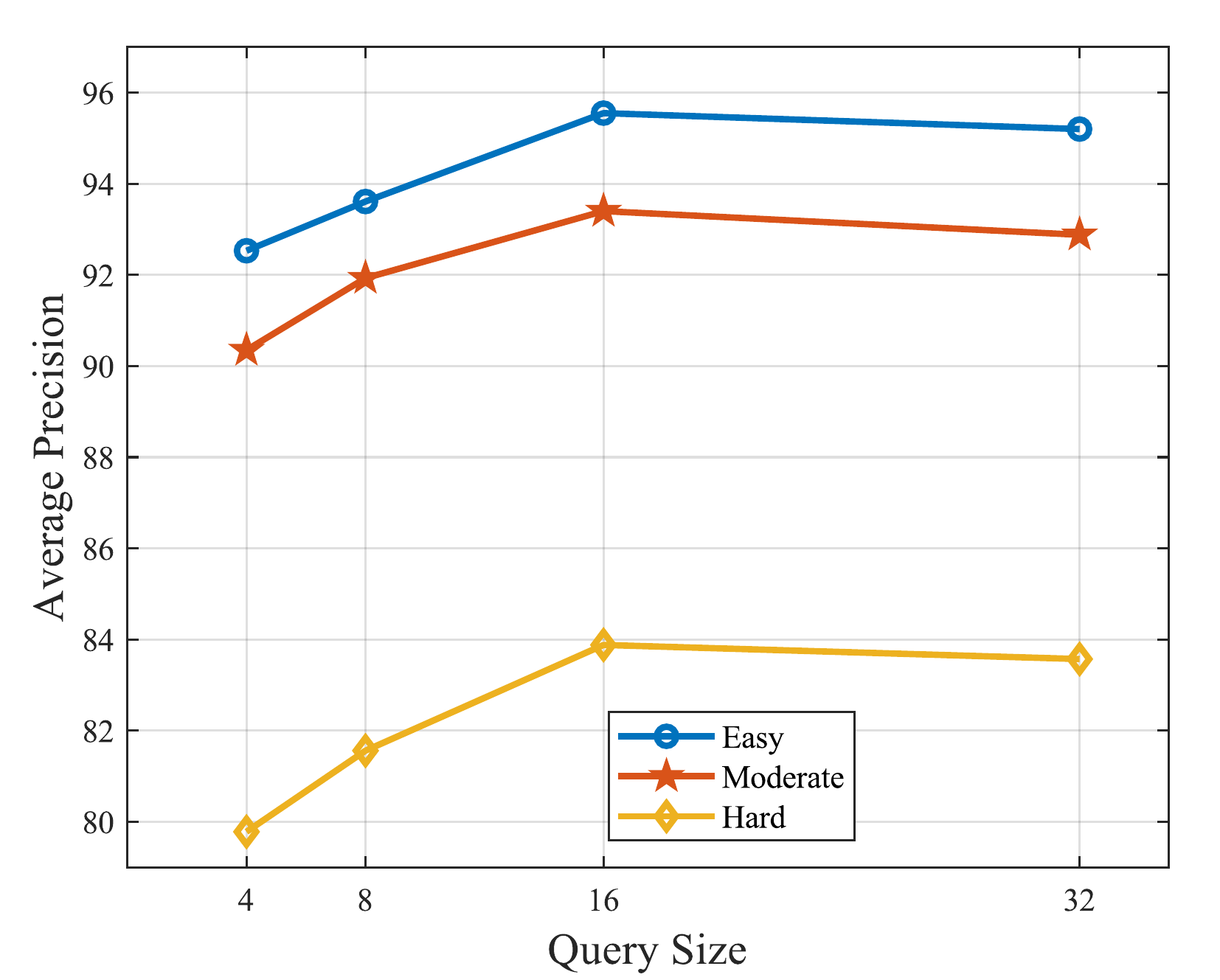}
    \label{vary_query_T_junction}
    } 
    \subfigure[]{
    \centering
    \includegraphics[scale=0.46]{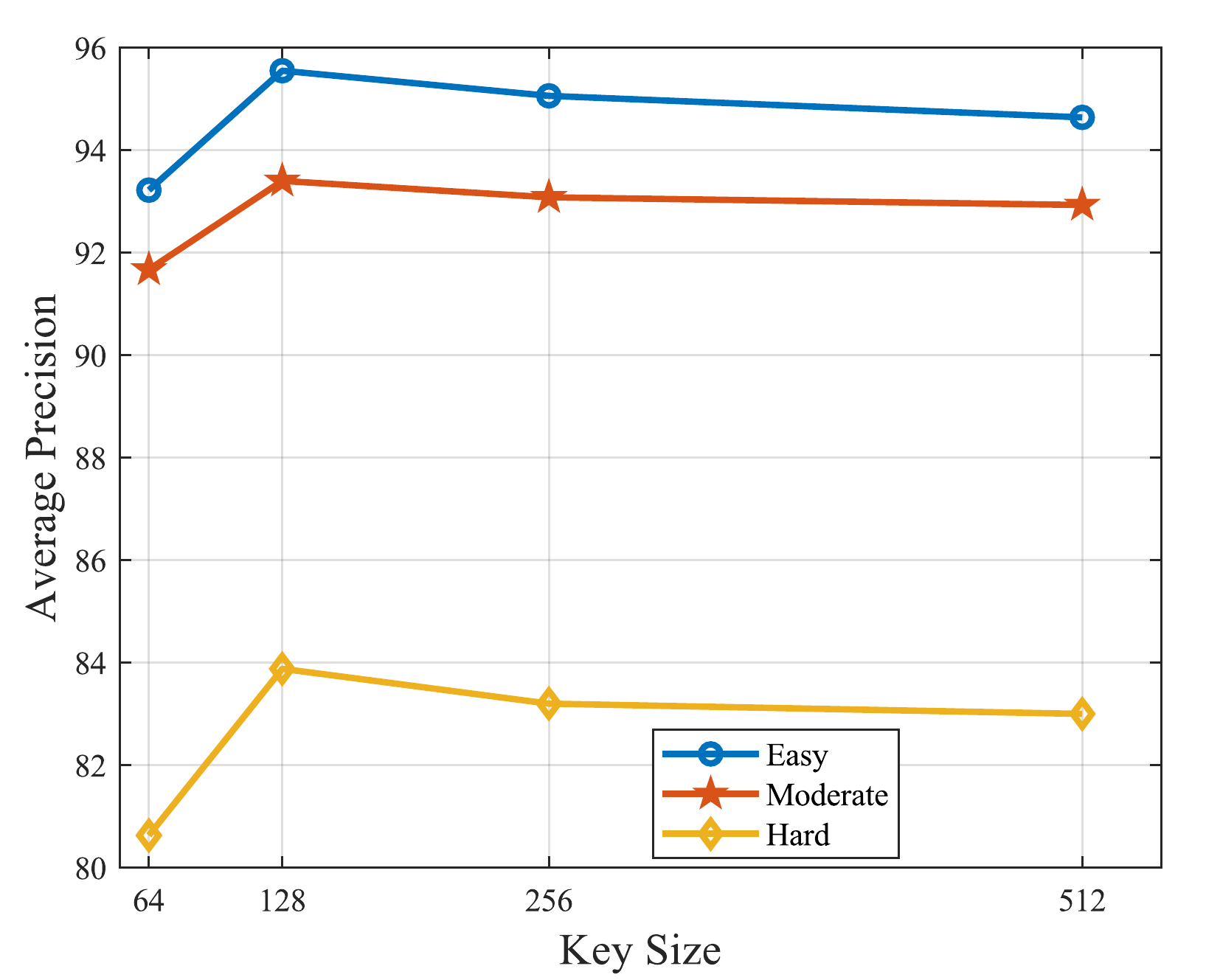}
    \label{vary_key_T_junction}
    }
    \caption{Ablation study on the size of query and keys under Easy, Moderate and Hard difficulties in T-junction scenario. (a) AP with key size of 128 versus query size from 4 to 32; (b) AP with query size of 32 versus key size from 64 to 512.}\label{query_and_key_T_junction}
\end{figure*}
\subsection{Experimental results}
\textbf{Impact of query and key sizes on detection performance} In our attention-based communication block, the sizes of query and key are considered to be different, where the size of query is much smaller than those of key. To determine the setup of the sizes of query and key, we do ablation studies on the different sizes of query and key for cooperative 3D object detection. Since we consider an $80.64m\times{71.68m}$ rectangle detection range with pillar size to $l_{x}=l_{y}=0.56$, the size of the pseudo-image obtained from the feature learning network is $64\times{128}\times{144}$. We first fix the key size as 128 and analyze the effect of query size on detection accuracy for AP by varying it from 4 to 32. Fig. \ref{vary_query_roundabout} and Fig. \ref{vary_query_T_junction} show detection accuracy of the proposed framework versus the number of query size under Easy Mod and Hard difficulties in roundabout and T-junction scenarios, respectively.  We observe that the detection accuracy of the framework is increased by increasing the number of query size under all difficulties. The detection accuracy is increased dramatically when the query size is increased from 4 to 16 and the performance improvement is flattening when the query size is bigger than 16. In addition, we conduct a similar experiment to study the effect of different key size of cooperative 3D detection. We fix the query size as 16 and show the effectiveness of different key sizes, in Fig. \ref{vary_key_roundabout} and Fig. \ref{vary_key_T_junction}. The experiment results show that increasing key size can improve detection accuracy and achieve the best performance at key size of 128. Once the key size is bigger than 128, the detection accuracy is decreasing and then flattening, since the key with size of 128 matches our feature maps with dimension of $64\times{128}\times(144)$. The experiment results indicate that a query size 16 paired with key size 128 can achieve amenable detection performance, we thus set query size to 16 and key size to 128 in our experiments.
\begin{figure*}[t]
    \centering
    {\centering
    \includegraphics[scale=0.30]{label_v3.png}
    }
    
    \subfigure[Locvehicle]{
    \centering
    \includegraphics[scale=0.50]{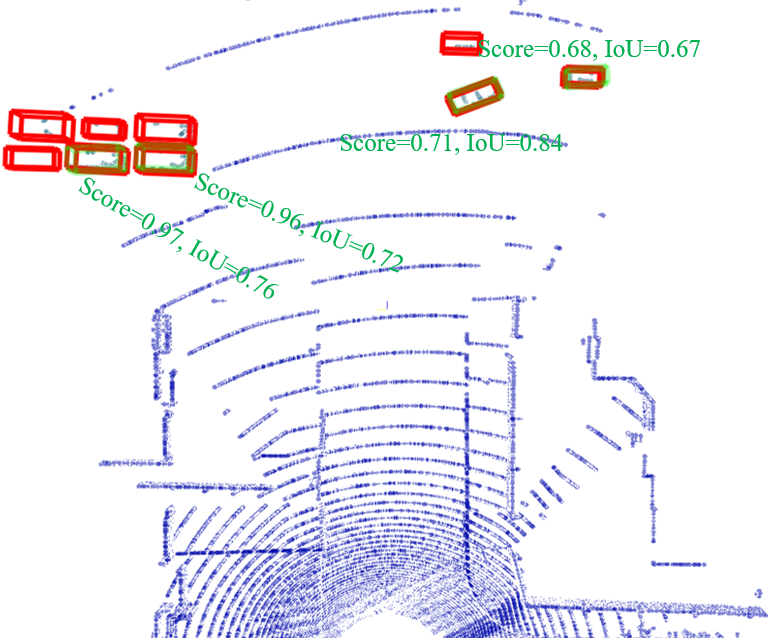}
    \label{LocaVehicle_T_junction}
    }
    \subfigure[RandSelect]{
    \centering
    \includegraphics[scale=0.50]{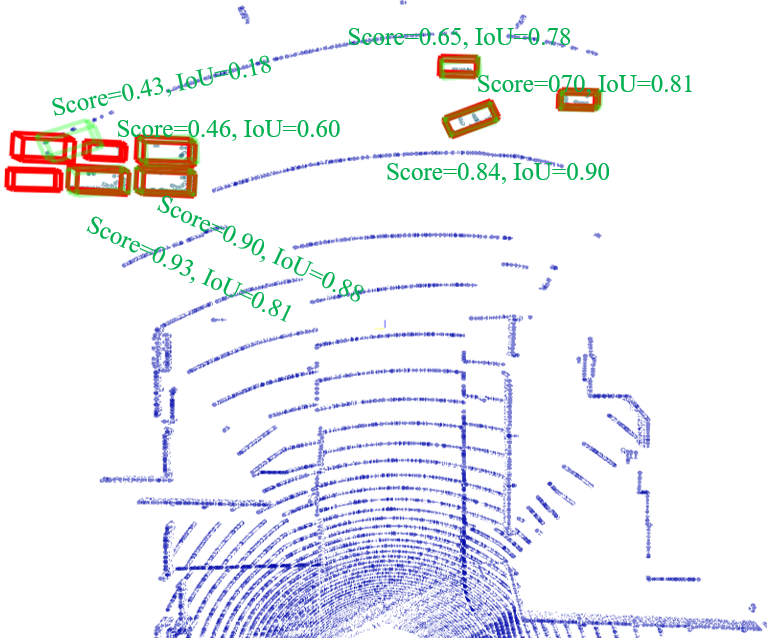}
    \label{RandSelect_T_junction}
    }
  
    \subfigure[CombineAll]{
    \centering
    \includegraphics[scale=0.50]{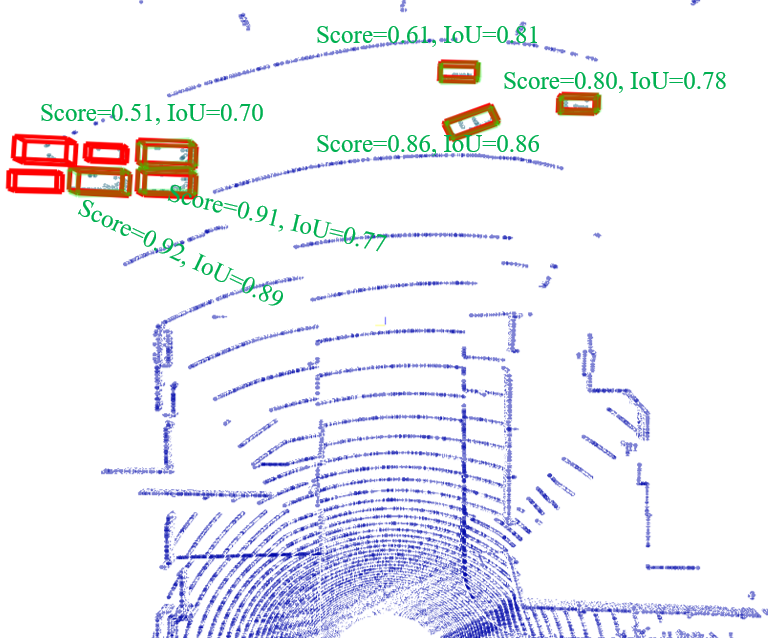}
    \label{Comball_T_junction}
    }
    \subfigure[Learn2com]{
    \centering
    \includegraphics[scale=0.50]{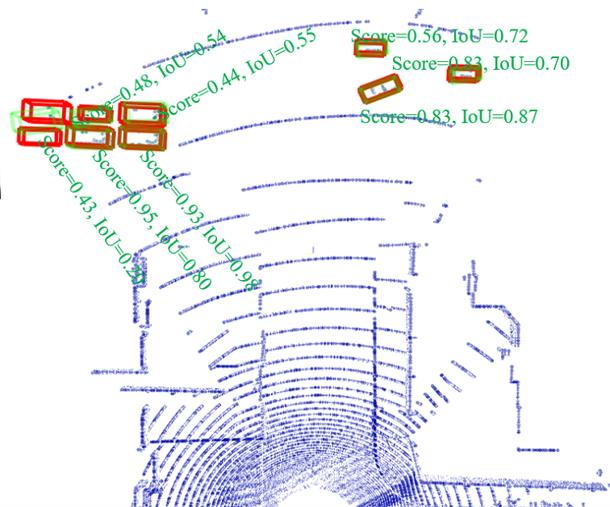}
    \label{learn2con_T_junction}
    }
    \caption{Detection examples of different detection schemes in T-junction scenario. Our method Learn2com performs better with higher detection accuracy. 
    }
    \label{T_junction_results} 
\end{figure*}
\begin{table*}[htbp]
\setlength{\tabcolsep}{2mm}
\centering
\caption{Detection performance comparison under CARLA-3D under near and far difficulties.}
\label{table:PERFORMANCE_detection_distance}
\begin{tabular}{|c|cccccc|clcccc|}
\hline
\multirow{3}{*}{} &
  \multicolumn{6}{c|}{roundabout AP@IoU 0.7} &
  \multicolumn{6}{c|}{T-junction AP@IoU 0.7} \\ \cline{2-13} 
 &
  \multicolumn{2}{c|}{mAP} &
  \multicolumn{2}{c|}{Car} &
  \multicolumn{2}{c|}{Truck} &
  \multicolumn{2}{c|}{mAP} &
  \multicolumn{2}{c|}{Car} &
  \multicolumn{2}{c|}{Truck} \\ \cline{2-13} 
 &
  \multicolumn{1}{c|}{Near} &
  \multicolumn{1}{c|}{Far} &
  \multicolumn{1}{c|}{Near} &
  \multicolumn{1}{c|}{Far} &
  \multicolumn{1}{c|}{Near} &
  Far &
  \multicolumn{1}{c|}{Near} &
  \multicolumn{1}{c|}{Far} &
  \multicolumn{1}{c|}{Near} &
  \multicolumn{1}{c|}{Far} &
  \multicolumn{1}{c|}{Near} &
  Far \\ \hline
LocVehicle &
  \multicolumn{1}{c|}{72.90} &
  \multicolumn{1}{l|}{70.77} &
  \multicolumn{1}{c|}{76.79} &
  \multicolumn{1}{c|}{61.06} &
  \multicolumn{1}{c|}{69.01} &
  80.48& 
  \multicolumn{1}{c|}{80.05} &
  \multicolumn{1}{c|}{62.79} &
  \multicolumn{1}{c|}{78.67} &
  \multicolumn{1}{c|}{51.67} &
  \multicolumn{1}{c|}{81.42} &
  73.91 \\ \hline
RandSelect &
  \multicolumn{1}{c|}{76.06} &
  \multicolumn{1}{l|}{73.53} &
  \multicolumn{1}{c|}{80.13} &
  \multicolumn{1}{c|}{64.95} &
  \multicolumn{1}{c|}{71.98} &
  82.11& 
  \multicolumn{1}{c|}{86.28} &
  \multicolumn{1}{c|}{66.24} &
  \multicolumn{1}{c|}{80.05} &
  \multicolumn{1}{c|}{56.06} &
  \multicolumn{1}{c|}{92.50} &
  76.42\\ \hline
CombALL &
  \multicolumn{1}{c|}{77.96} &
  \multicolumn{1}{l|}{75.74} &
  \multicolumn{1}{c|}{81.65} &
  \multicolumn{1}{c|}{67.67} &
  \multicolumn{1}{c|}{74.26} &
  83.81&
  \multicolumn{1}{c|}{86.06} &
  \multicolumn{1}{c|}{65.72} &
  \multicolumn{1}{c|}{82.12} &
  \multicolumn{1}{c|}{56.18} &
  \multicolumn{1}{c|}{90.00} &
  75.25 \\ \hline
Learn2com &
  \multicolumn{1}{c|}{\textbf{79.08}} &
  \multicolumn{1}{l|}{\textbf{76.42}} &
  \multicolumn{1}{c|}{\textbf{82.75}} &
  \multicolumn{1}{c|}{\textbf{68.86}} &
  \multicolumn{1}{c|}{\textbf{75.40}} &
  \textbf{83.98}& 
  \multicolumn{1}{c|}{\textbf{90.85}} &
  \multicolumn{1}{c|}{\textbf{66.66}} &
  \multicolumn{1}{c|}{\textbf{86.70}} &
  \multicolumn{1}{c|}{\textbf{56.43}} &
  \multicolumn{1}{c|}{\textbf{95.00}} &
  \textbf{76.83} \\ \hline
\end{tabular}
\end{table*}  

\textbf{Detection performance comparison under Easy, Moderate and Hard difficulties}
Detection results in Table \ref{table:PERFORMANCE_detection_kitti} show that the proposed Learn2com model performs better than all baseline models under all difficulties. In addition, all communication-based models have higher detection accuracy than the local processing method LocVehicle under all difficulties in both scenarios.
Specifically, in the roundabout driving scenario, Learn2com gets around 4.56 higher mAP (mean average precision) than LocVehicle, 0.75 higher mAP than RandSelect, and 1.61 higher mAP than CombAll under Moderate difficulty level. 
In the T-junction scenario, compared to LocVehicle, Learn2com gets around 5.37 higher mAP under Moderate.
Moreover, the mAP of Learn2com is slightly higher (0.88) than those of CombAll and (2.13) than those of RandSelect under Moderate.
Fig. \ref{roundabout_results} shows detection examples of different detection schemes in the roundabout scenario as shown in Fig. \ref{roundabout_scene}. Fig. \ref{T_junction_results} shows detection examples of different detection schemes in the T-junction scenario given in Fig. \ref{T_junction_scene}.

\textbf{Detection accuracy comparison under near and far difficulties} We define the objects within 20 meters far away from the center of the vehicle as ``near", and those beyond 20 meters as ``far" objects. As shown in Table \ref{table:PERFORMANCE_detection_distance}, Learn2com outperforms all baseline models and all communication-based models perform better than local processing model LocVehicle. 
Compared to LocVehicle method, Learn2com gets around 6.18 and 5.65 higher mAP under near and far difficulties in the roundabout scenario, respectively, and gets 10.80 and 3.87 higher mAP under near and far difficulties in the T-junction scenario, respectively. 
Moreover, the performance of Learn2com is better than CombAll and RandSelect.

\begin{table}[H]
    \centering
    \caption{Comparison results of bandwidth usage and AIB on CARLA-3D. We calculate the AIB under Moderate difficulty level.}
    \label{table:PERFORMANCE_bandwidth}
    \begin{tabular}{|c|c |c| c |c|}
    \hline
    &\multicolumn{2}{c|}{roundabout} & \multicolumn{2}{c|}{T-junction}
    \\ \cline{2-5} 
    & Bandwidth (kbpf) & AIB & Bandwidth (kbpf) & AIB \\
    \hline
    RandSelect & \textbf{4608} & 0.85 & \textbf{4608} & 0.72 \\ 
    \hline
    CombAll & 13824 & 0.22 &  9216 & 0.50 \\ 
    \hline
    \textbf{Learn2com} & \textbf{4608.06} & \textbf{1.01} & \textbf{4608.06} & \textbf{1.19} \\ 
    \hline
    \end{tabular}
\end{table}
\textbf{Bandwidth usage comparison} We compare the bandwidth usage between our method and other two communication-based baseline methods, which are CombAll and RandSelect.
LocVehicle uses local information for detection and has no bandwidth consumption. The comparison results are shown in Table \ref{table:PERFORMANCE_bandwidth}. Our method achieves highest AIB score and uses much less bandwidth than CombAll, and a bit more bandwidth than RandSelect. This can be explained that CombAll gathers learned feature maps from all neighboring infrastructures, while our method broadcasts query to all infrastructures and only collects feature maps from one neighboring infrastructure.
In our experiments, the query size (16) is much smaller than the feature map size ($64\times{128}\times{144}$ per infrastructure). In addition, compared to our method, RandSelect model directly selects one infrastructure without broadcasting query information, and thus, the bandwidth used to broadcast query will be saved.

\section{Conclusion}
\label{con}
In this work, we proposed a novel cooperative perception framework (Learn2com) for 3D object detection using an attention-based communication scheme. Our work is based on the fact that an autonomous vehicle can perceive the driving environment better by combining sensing information from neighboring infrastructures. The proposed framework consists of three modules, which are feature learning networks, an attention-based communication block, and a region proposal network. It first maps point clouds into pseudo-images using the feature learning networks and then uses the communication block to learn effective communication schemes and reduce bandwidth usage by transmitting fewer data. After that, the region proposal network produces object classification results and 3D bounding boxes. The proposed framework was trained in an end-to-end manner, and a centralized communication scheme was adopted during training and distributed communication scheme was executed during inference. 

To evaluate the effectiveness of the proposed framework on cooperative 3D object detection from point clouds, we designed a new dataset CARLA-3D, including a roundabout and a T-junction driving scenario, using an open-source urban driving simulator Car Learning to Act (CARLA)\ \cite{CARLA}. The proposed cooperative perception framework was analyzed and compared with three baseline models. Experiment results showed that the proposed framework with attention mechanism achieves better detection performance than the centralized perception method (CombAll), and consumes much fewer communication costs in terms of transmitted bits and accuracy-improvement-to-bandwidth-usage. In addition, Learn2com gets better 3D object detection performance than the local process model (LocVehicle) and the random selection model (RandSelect), since the communication block learns to fuse feature information from one neighboring infrastructure with a general attention mechanism. This work gave an extensive study on saving bandwidth for collaborative 3D object detection by learning to communicate with neighboring infrastructures, which has significant potential in practical autonomous driving systems. The key idea of this work is to use point cloud data for collaborative 3D object detection, and save bandwidth via communicating with one neighboring sensor per time frame. For future research, it will be interesting to study how to design a communication trigger mechanism to further save bandwidth and incorporate RGB-D image data to further improve collaborative perception accuracy.

\end{document}